\pdfoutput=1
 \documentclass[final,3p,times]{elsarticle}
\usepackage{amssymb}
\usepackage{hyperref}
\usepackage{color}

\usepackage{lineno}
\usepackage{url}
\usepackage{subfigure}
\usepackage{epstopdf}
\newcommand{\etal}{\textit{et al}}

\usepackage{amsmath}
\DeclareMathOperator*{\argmax}{argmax}

\journal{ISPRS Journal of Photogrammetry and Remote Sensing, in press.}

\begin{document}
	\begin{frontmatter}
		
		\title{Building Instance Classification Using Street View Images}
		
		
		\author[firstaddress]{Jian Kang}
		
		\author[secondaryaddress]{Marco K\"orner}
		\author[firstaddress]{Yuanyuan Wang}
		\author[thirdaddress]{Hannes Taubenb\"ock}
		\author[firstaddress,fourthaddress]{Xiao Xiang  Zhu\corref{mycorrespondingauthor}}
		\cortext[mycorrespondingauthor]{Corresponding author (Email:xiao.zhu@dlr.de)}
		
		\address[firstaddress]{Signal Processing in Earth Observation (SiPEO), Technical University of Munich (TUM), 80333 Munich, Germany}
		\address[secondaryaddress]{Chair of Remote Sensing Technology, Technical University of Munich (TUM), 80333 Munich, Germany}
		\address[thirdaddress]{German Remote Sensing Data Center (DFD) (IMF), German Aerospace Center (DLR), 82234 Wessling, Germany}
		\address[fourthaddress]{Remote Sensing Technology Institute (IMF), German Aerospace Center (DLR), 82234 Wessling, Germany}
		
		\begin{abstract}
			\textit{This is the pre-print version, to read the final version please go to ISPRS Journal of Photogrammetry and Remote Sensing, Elsevier. (https://doi.org/DOI: 10.1016/j.isprsjprs.2018.02.006)}. Land-use classification based on spaceborne or aerial remote sensing images has been extensively studied over the past decades. Such classification is usually a patch-wise or pixel-wise labeling over the whole image. But for many applications, such as urban population density mapping or urban utility planning, a classification map based on individual buildings is much more informative. However, such semantic classification still poses some fundamental challenges, for example,  how to retrieve fine boundaries of individual buildings. In this paper, we proposed a general framework for classifying the functionality of individual buildings. The proposed method is based on Convolutional Neural Networks (CNNs) which classify fa\c{c}ade structures from street view images, such as Google StreetView, in addition to remote sensing images which usually only show roof structures. Geographic information was utilized to mask out individual buildings, and to associate the corresponding street view images. We created a benchmark dataset which was used for training and evaluating CNNs. In addition, the method was applied to generate building classification maps on both region and city scales of several cities in Canada and the US.
		\end{abstract}
		
		
	\end{frontmatter}
	
%	\linenumbers
	
	\section{Introduction}
	The classification of land cover from Earth Observation (EO) images in complex urban environments has been a focus in remote sensing over the past decades \cite{anderson1976land,pal2003assessment,yuan2005land,stefanov2001monitoring,rodriguez2012assessment,albert2017using}. Beyond, high resolution spaceborne and aerial images are one of the handful information sources for monitoring urban development on large scales. 
	
\begin{figure}
		\centering
		\subfigure[commercial]{\label{subfig_commercial} \includegraphics[width=0.3\textwidth]{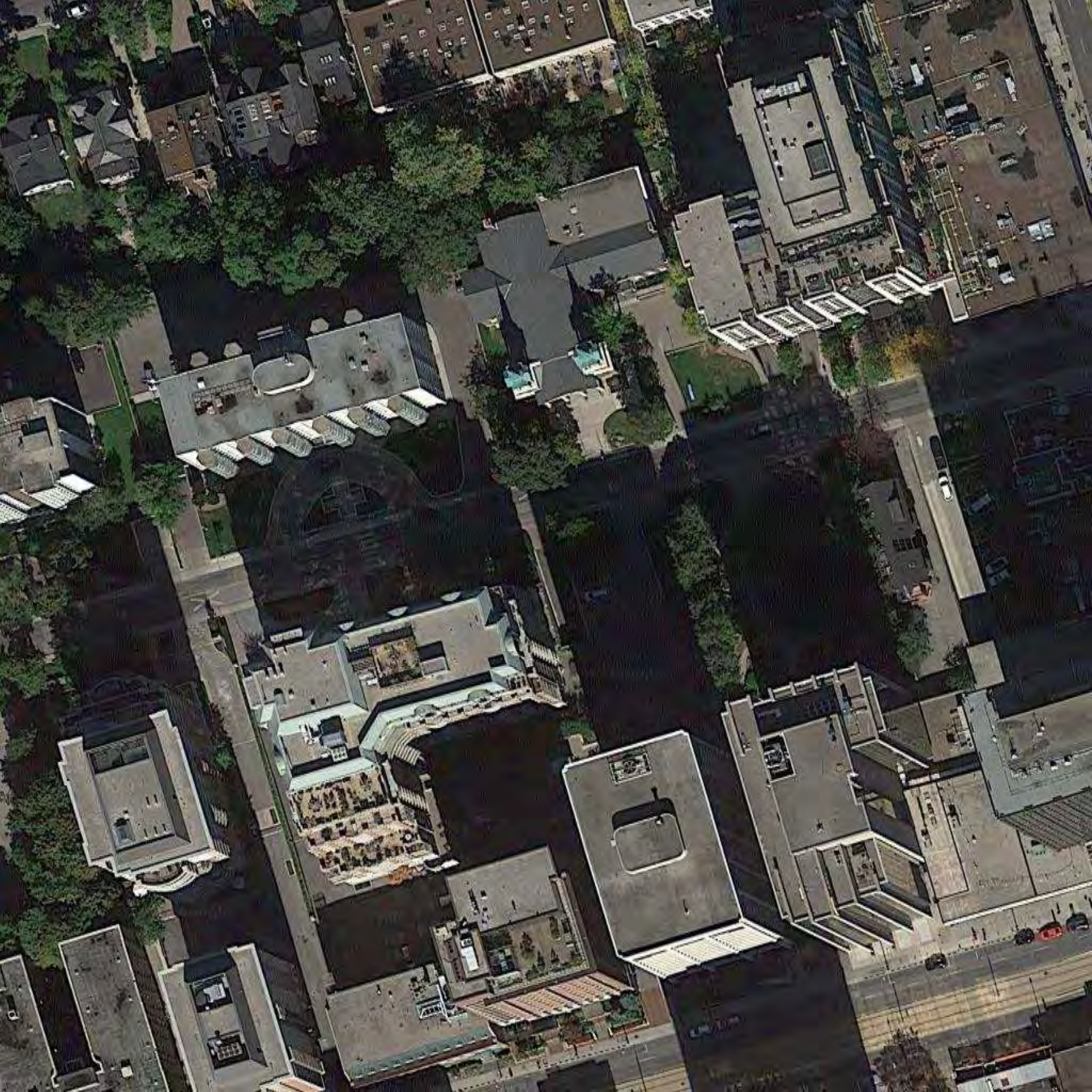}}
		\subfigure[residential]{\label{subfig_residential} \includegraphics[width=0.3\textwidth]{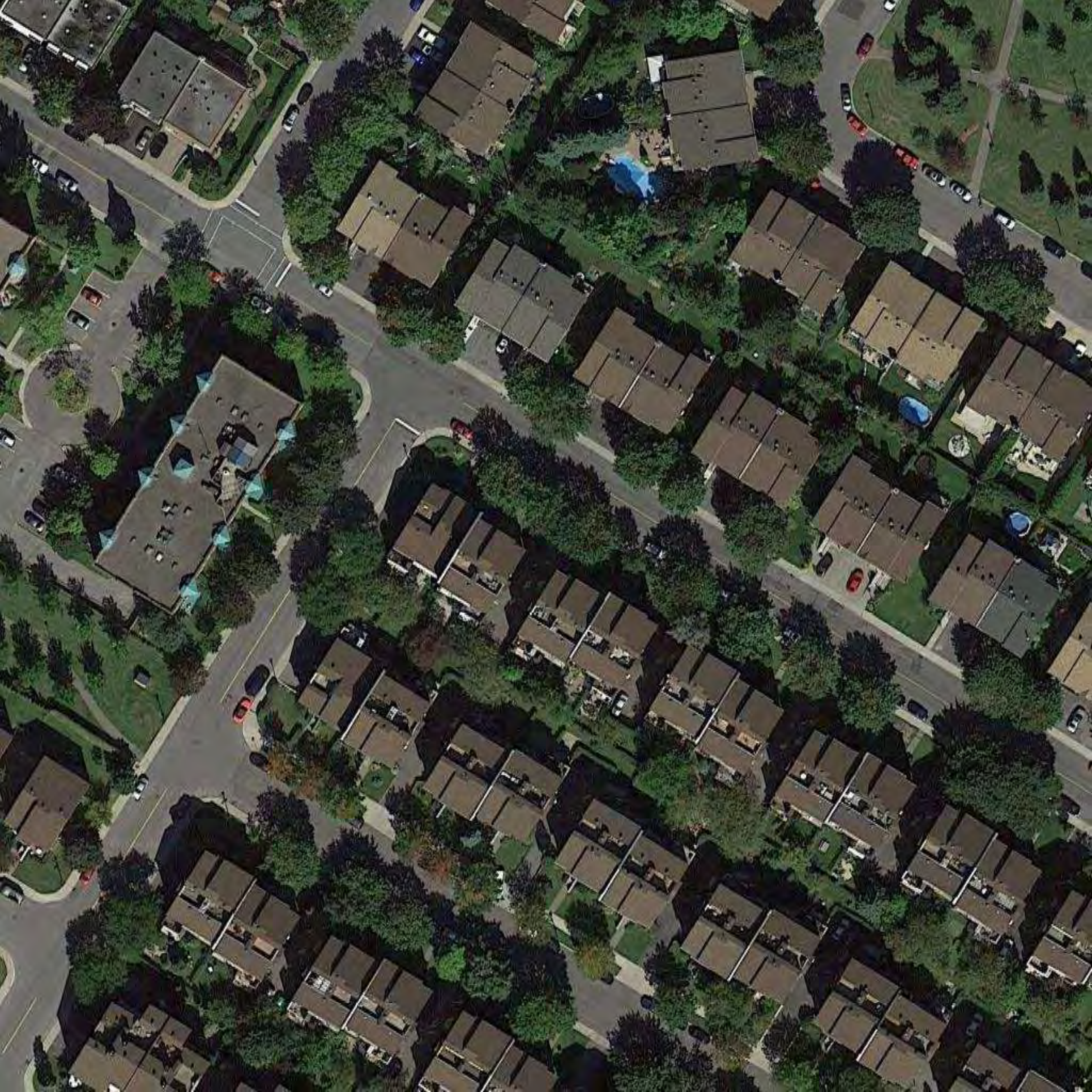}}
		\subfigure[industrial]{\label{subfig_industrial} \includegraphics[width=0.3\textwidth]{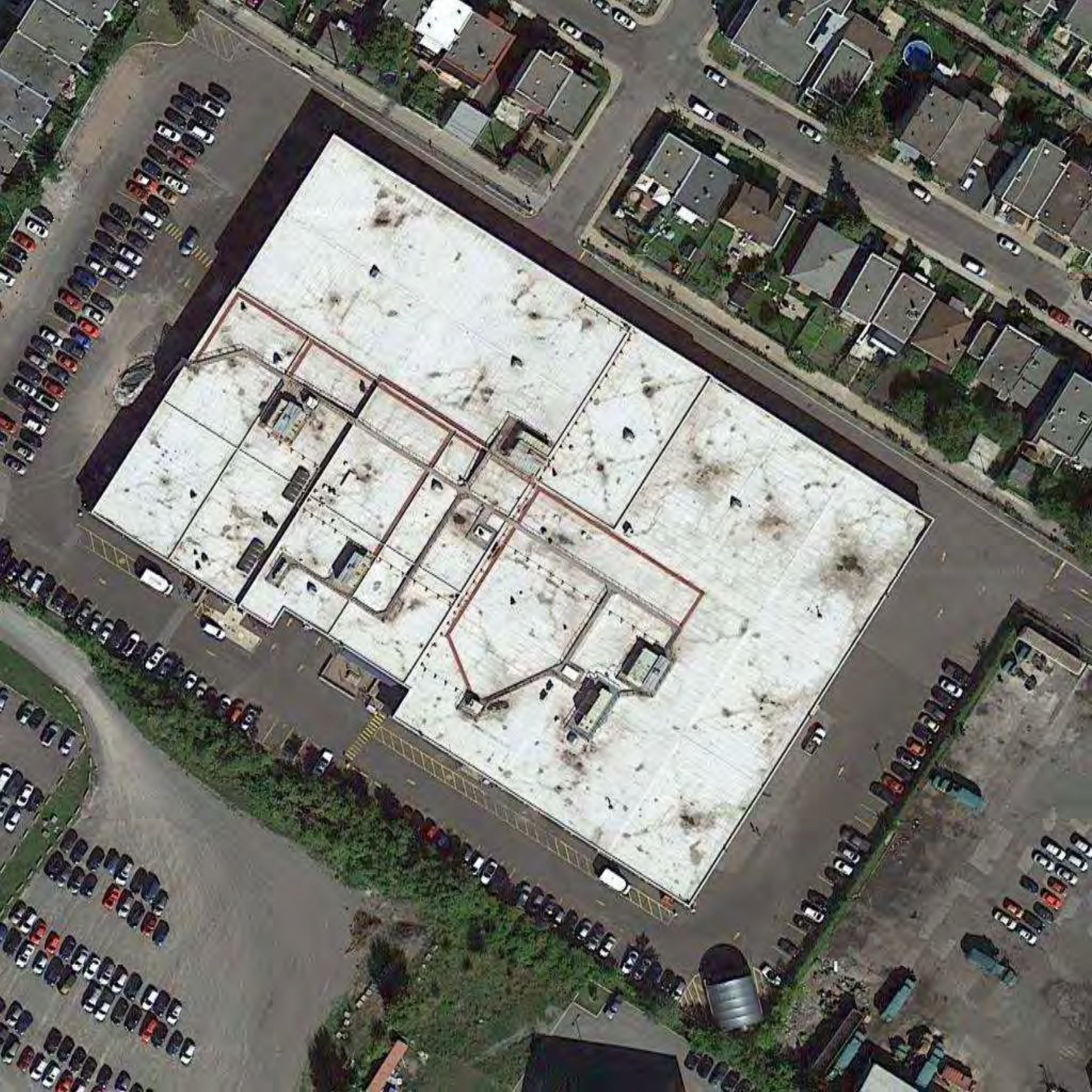}}
		\caption{Examples of land-use classification}
		\label{fg:land_use_examples}
	\end{figure}

\begin{figure}
		\centering
		\includegraphics[width=0.8\textwidth]{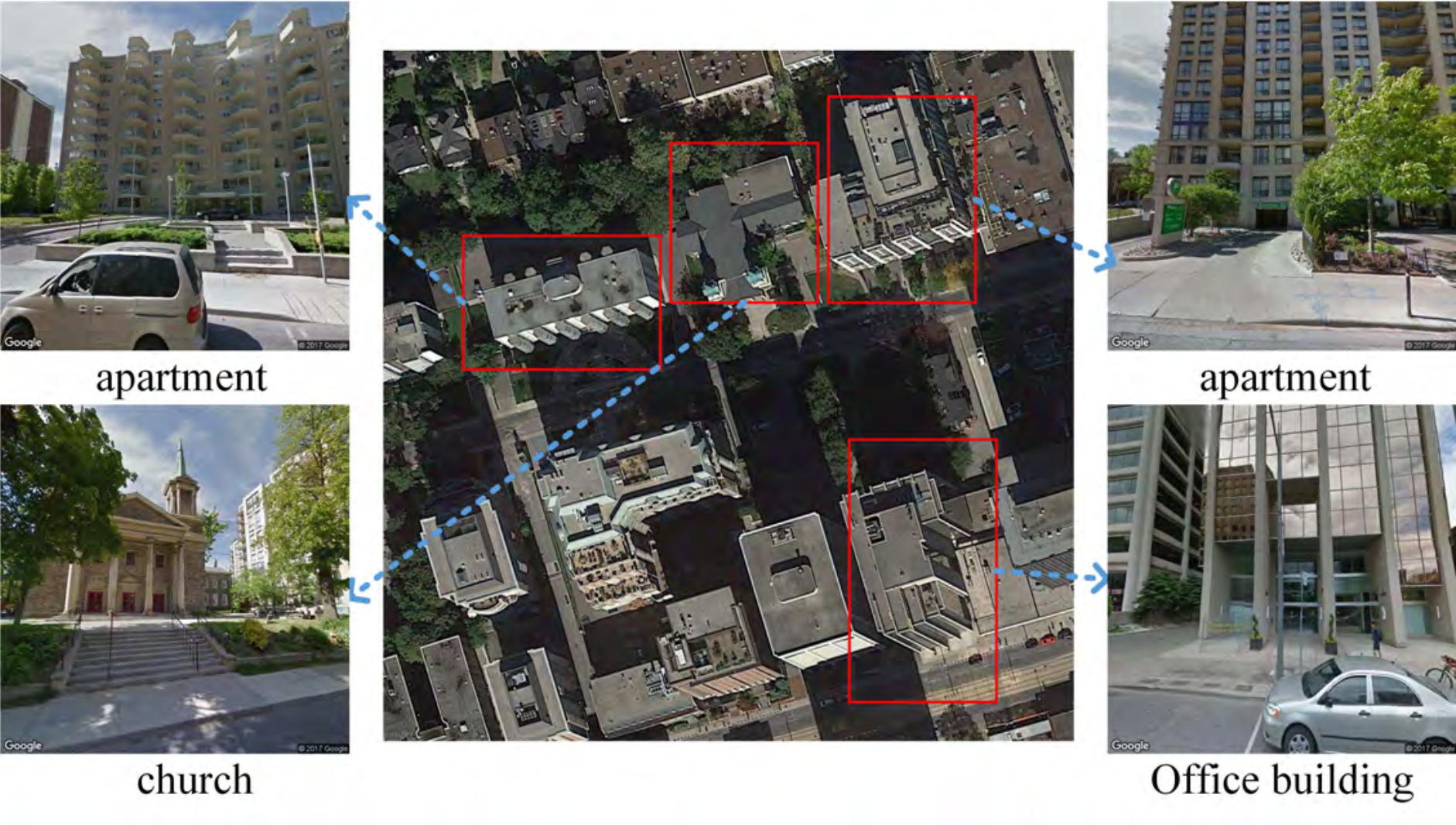}
		\caption{The commercial land-use area as shown in Figure \ref{fg:land_use_examples} (a), along with the street view images for some buildings selected by the red rectangles. These buildings do not belong to the same category, even though they are located in the same land-use area. Besides, compared to the roof structures, the information of fa\c{c}ade structures displayed in street view images is richer and more sufficient to be used for building instance classification.}
		\label{fg:land_use_GSV_examples}
	\end{figure}

	However, the transfer from land cover to land use in EO-data is complex and relies mostly on the geometry and the appearance of individual buildings and the patterns they group together \cite{lu2006use,gong1992comparison,paola1995detailed,pacifici2009neural,khorram1987comparson,di2000land,cheng2015effective,huang2017multi,huang2014multi,huang2015spatiotemporal}. The correlation of physical indicators such as building volumes, density or alignment has been used to infer the usage of buildings, e.g. as commercial areas (e.g. Figure \ref{fg:land_use_examples}(a)), residential areas (e.g. Figure \ref{fg:land_use_examples}(b)) or industrial areas (e.g. Figure \ref{fg:land_use_examples}(c)). Nevertheless, such pattern analysis can not be directly transferable to the classification of individual buildings as we go to a finer level of urban intrinsic scale. For example, Figure \ref{fg:land_use_examples}(a) shows a commercial area comprised of multiple high-rise buildings. However, the label "commercial area" cannot be assigned to all the building instances within it. As illustrated in Figure \ref{fg:land_use_GSV_examples}, the corresponding street view images show that the commercial area is comprised of a few apartments, office buildings, and one church. This also applies to the example shown in Figure \ref{fg:land_use_examples} (b) and (c), where both the residential and industrial areas are comprised of buildings with different functionalities.
	As can be seen, land-use classification at a level of individual buildings is not a trivial task. Usually, such a classification map is only obtainable through city cadastral databases, not accessible or sometimes even not existent. Updating such databases without automatic methods can be very labor intensive. Hence, automatically achieving a building instance-level classification is necessary and can be beneficial for applications related with urban planning. 
	Towards an automatic classification of individual buildings, the challenges are twofold. Firstly, remote sensing images usually only contain roof structures due to their nadir-looking imaging geometry. The visual difference of the roofs between certain building classes, e.g. apartments and office buildings, can be subtle, as an example shown in Figure \ref{fg:land_use_GSV_examples}. Secondly, the extraction of building footprints directly from remote sensing images is still under preliminary research. A clear segmentation of building footprints usually requires height information which comes at an additional cost. 	
	In this paper, we propose a general framework to tackle the abovementioned challenges, which exploits the information extraction from freely available street view images and online geographic maps. Specifically, fa\c{c}ade structures shown in online street view images are sufficiently rich for building functionality classification, and the online map services, such as OpenStreetMap \cite{OpenStreetMap} or Google Maps, can provide the building footprints which can be associated to street view images via their geographic locations. As shown in Figure \ref{fg:land_use_GSV_examples}, the fa\c{c}ades displayed in street view images reveal much more details of different types of buildings than the corresponding roof patches. Therefore, building instances are classified based on their geo-tagged street view images in the proposed method, and the inferred labels are then linked to individual building footprints through spatial clustering. We also build a benchmark dataset of building street view images to train Convolutional Neural Networks (CNNs) for the classification over large areas, as CNN has been demonstrated its powerful ability in the tasks of this sort \cite{russakovsky2015imagenet,zhou2016places,lin2014microsoft}. 
	
	In a summary, the contributions of this paper are listed as follows:
		\begin{itemize}
			\item Proposed a general framework for land-use classification at a level of individual buildings.
			\item Built a street view benchmark dataset for training building instance CNN classifiers based on fa\c{c}ade structures. The dataset utilized in this paper can be downloaded via \url{www.sipeo.bgu.tum.de/downloads/BIC_GSV.tar.gz}
			\item The obtained building classification maps demonstrated the potentials for many innovative urban analysis, e.g. very high resolution urban population density mapping, urban social structure understanding, city economy structure analysis and general urban planning.
	\end{itemize}

	\section{Related work}
	Feature extraction from remote sensing images plays a vital role in land-use classification. Handcrafting features have been well studied for decades, such as scale-invariant feature transform (SIFT) \cite{lowe1999object} encoded by bag of visual words (BoVW) \cite{yang2010bag,cheriyadat2014unsupervised,zhu2016bag}, multiple textural features \cite{xu2016multiple}, 3D features derived from a digital surface model \cite{taubenbock2013delineation} and features learned by sparse coding methods \cite{wang2014spatial,yang2014data,sun2015task,rigas2013low,zhang2015saliency,tuia2016nonconvex,yao2016semantic,cheng2015effective,tuia2015multiclass}.
	
	Recently, many approaches based on deep learning techniques have emerged \cite{cheng2017remote,ma2016semisupervised,zhang2017hybrid}. Chen \etal. \cite{chen2014deep} proposes a hierarchical feature extraction method via stacked autoencoders, which merges both spectral and spatial information of hyperspectral images for land-use classification. In \cite{zou2015deep}, deep belief networks are employed for the feature learning in remote sensing scene classification.  Both \cite{penatti2015deep} and \cite{marmanis2016deep} investigate the possibility of transferring features learned by CNN from \textit{ImageNet} dataset \cite{deng2009imagenet} to achieve remote sensing image classification by fine-tuning procedures. To improve the composition-based inference of land-use classes, multiscale CNN-based approaches are developed in \cite{zhao2016learning,luus2015multiview,liu2016adaptive}. By exploiting deep Boltzmann machine, a novel weakly supervised learning approach for object detection in remote sensing images is introduced \cite{han2015object}. For effectively dealing with the problem of object rotation variations, a rotation-invariant CNN model is proposed in \cite{cheng2016learning}. Based on greedy layerwise unsupervised pretraining, \cite{romero2016unsupervised} proposes a novel unsupervised deep feature extraction method. Taking advantage of geographical information from OpenStreetMap, a fully convolutional neural network is trained to achieve pixel-wise classifications in optical images on large scales \cite{maggiori2017convolutional}. Recurrent Neural Network (RNN) is also proved to be efficient for classifying sequence-based data like hyperspectral images \cite{7914752}. An end-to-end fully Conv-Deconv network for unsupervised spectral-spatial feature extraction in hyperspectral images has been proposed in \cite{mou2018deepresidual}. In order to better interpret land-uses of Synthetic Aperture Radar (SAR) images in urban areas, \cite{hughes2018identifying} proposes a pseudo-siamese CNN for identifying corresponding patches in very-high-resolution (VHR) optical and  SAR remote sensing imagery. Surveys about the applications of deep learning techniques to land-use classification with remote sensing images are proposed in \cite{zhang2016deep,zhu2017deeplearning}. 
	
	Even the abovementioned literature is of course not exhaustive, none of them have explicitly addressed the land-use classification at a level of individual buildings.

	\begin{figure}
		\centering
		\includegraphics[width=\textwidth]{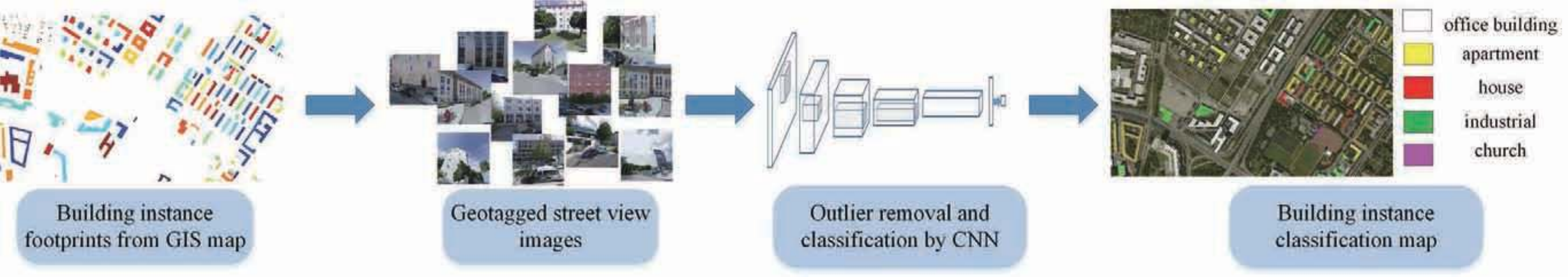}
		\caption{The proposed workflow for land-use classification at a level of individual buildings.}
		\label{fig_workflow}
	\end{figure}
	
	\section{Overall workflow}
	As illustrated in Figure \ref{fig_workflow}, the proposed workflow for building instance classification contains the following steps:
	\begin{itemize}
		\item Retrieval of building footprints and associated street view images.
		\item Outlier removal by the pretrained CNN on Places2 dataset \cite{zhou2016places}. 
		\item Building instance classification by the CNN trained on our benchmark dataset.
	\end{itemize}
	
	\subsection{Retrieval of building footprints and street view images}
\begin{figure}
		\centering
		\includegraphics[width=0.8\textwidth]{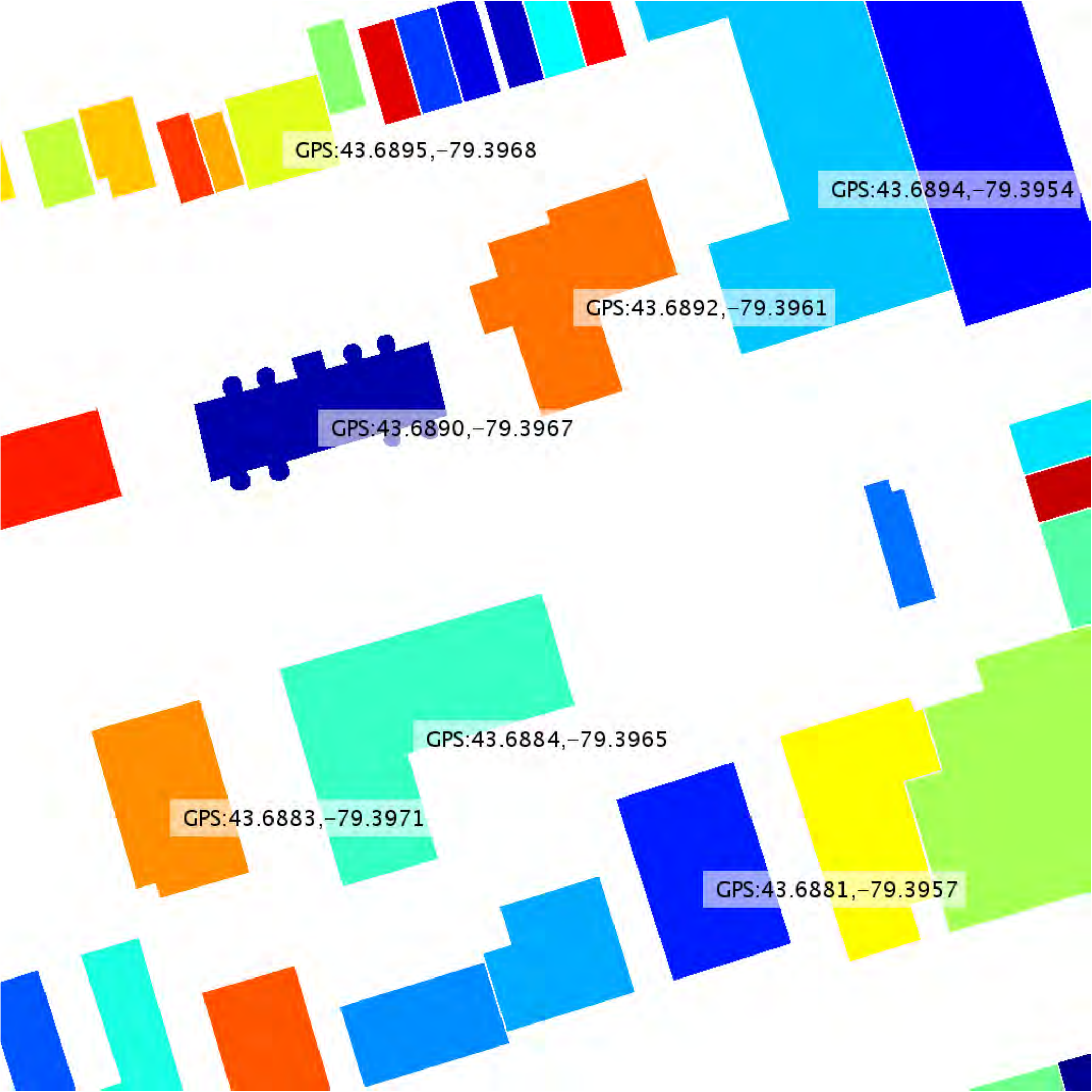}
		\caption{Geographic information (GPS) retrieved from Google Maps of the remote sensing image in Figure \ref{fg:land_use_GSV_examples}, with the color randomly assigned to each building mask.}
		\label{fg:land_use_GSV_examples_GIS_infor}
	\end{figure}

	The building footprints and their geographic locations, can be retrieved from online geographic information systems (GIS), such as OpenStreetMap or Google Maps. For example, the building footprints of the area shown in Figure \ref{fg:land_use_GSV_examples} are displayed in Figure \ref{fg:land_use_GSV_examples_GIS_infor}, along with the associated GPS coordinates (latitude, longitude). The color is randomly assigned to indicate different building instances. Given these GPS coordinates, we can download the corresponding Google StreetView images \cite{anguelov2010google}  which show fa\c{c}ade structures of individual buildings, since the retrieved images can display these specific locations by the closest panoramas.   
	
	\subsection{Outlier removal by pretrained CNN on Places2 dataset}
\begin{figure}
		\centering
		\includegraphics[width=0.8\textwidth]{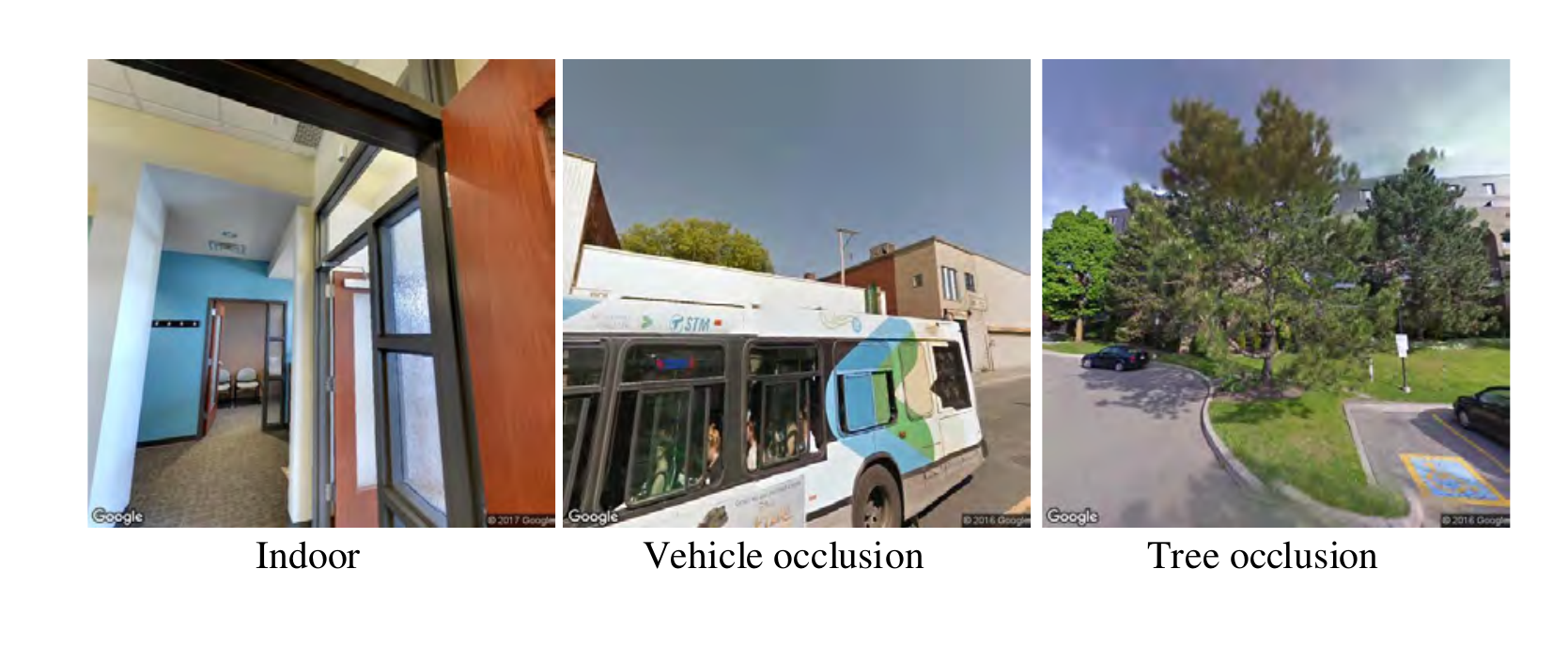}
		\caption{Outlier examples of the retrieved street view images. We can see that there is no available information of building fa\c{c}ades for the classification.}
		\label{fg:outlier_example}
	\end{figure}

	Due to the uncontrolled quality of street view images, many of them cannot be directly utilized for the building classification. For example, as shown in Figure \ref{fg:outlier_example}, one retrieved image is taken from the  building interior and the other two buildings are occluded by a vehicle and trees on the side-walks. Therefore, the corresponding fa\c{c}ade structures are not available for classifying these buildings. These outliers can severely influence the classification results. 
	For removing them, we employ the released VGG16 model \cite{simonyan2014very} trained on Places2 dataset \cite{zhou2016places} to preliminarily screen the street view images, as this architecture has achieved the highest top-1 accuracy\footnote{\url{https://github.com/metalbubble/places365}}. The dataset contains almost 10 million scene photos, labeled with 476 scene categories and attributes, which include the building-related categories, i.e. [\textit{apartment, church, house, industrial area, museum, building facade, embassy, hospital, parking garage, hotel}]. Only the images belonging to the abovementioned categories are preserved for the follow-up classification. 
	
	\subsection{Building instance classification}
	\begin{table}
		\caption{Building class descriptions from OpenStreetMap}
		\centering
		\renewcommand{\arraystretch}{1.3}
		\begin{tabular}{l | l}
			\hline
			apartment & \parbox[t][][t]{9cm}{A building arranged into individual dwellings, often on separate floors. May also have retail outlets on the ground floor.}\\
			church & A building that was built as a church. \\
			garage & \parbox[t][][t]{9cm}{A building suitable for the storage of one or possibly more motor vehicle or similar.}  \\
			house & \parbox[t][][t]{9cm}{A dwelling unit inhabited by a single household (a family or small group sharing facilities such as a kitchen).} \\
			industrial & \parbox[t][][t]{9cm}{A building where some industrial process takes place.}\\
			office building & \parbox[t][][t]{9cm}{A building where non-specific commercial activities take place.}\\
			retail & \parbox[t][][t]{9cm}{A building primarily used for selling goods that are sold to the public.}\\
			roof & \parbox[t][][t]{9cm}{A structure that consists of a roof with open sides, such as a rain shelter, and also gas stations.}\\
			\hline
		\end{tabular}
		\label{tb:building_cls_description}
	\end{table}
	
\begin{figure}
		\centering
		\includegraphics[width=\textwidth]{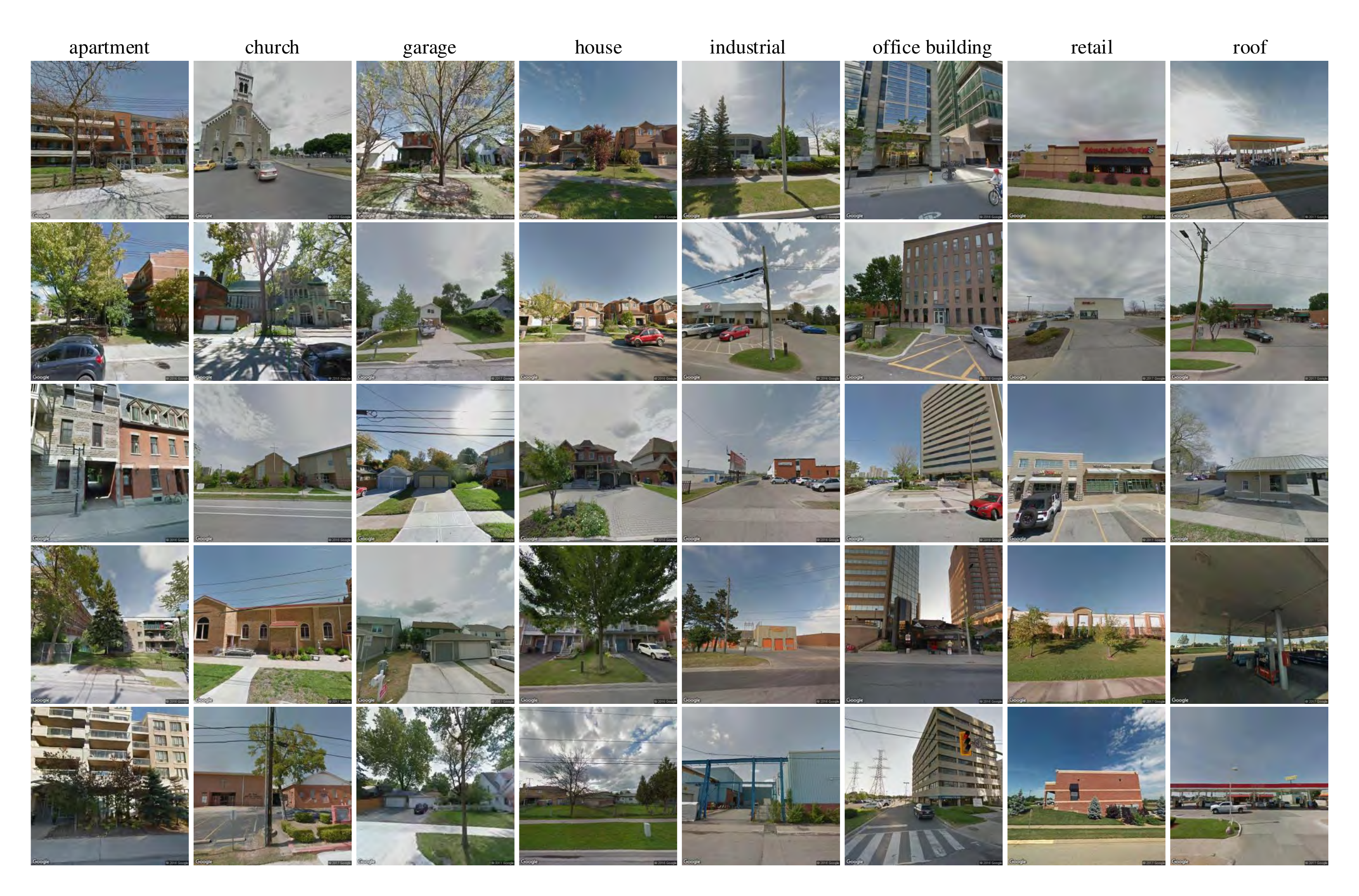}
		\caption{Examples of the benchmark dataset. It totally contains 19658 street view images of buildings with eight classes, i.e. \textit{apartment, church, garage, house, industrial, office building, retail and roof}. The images are downloaded from Google StreetView \cite{anguelov2010google}, and the associated labels are jointly retrieved from OpenStreetMap based on the geographic information.}
		\label{fg:examples_dataset}
	\end{figure}

\begin{figure}
		\centering
		\includegraphics[width=0.5\textwidth]{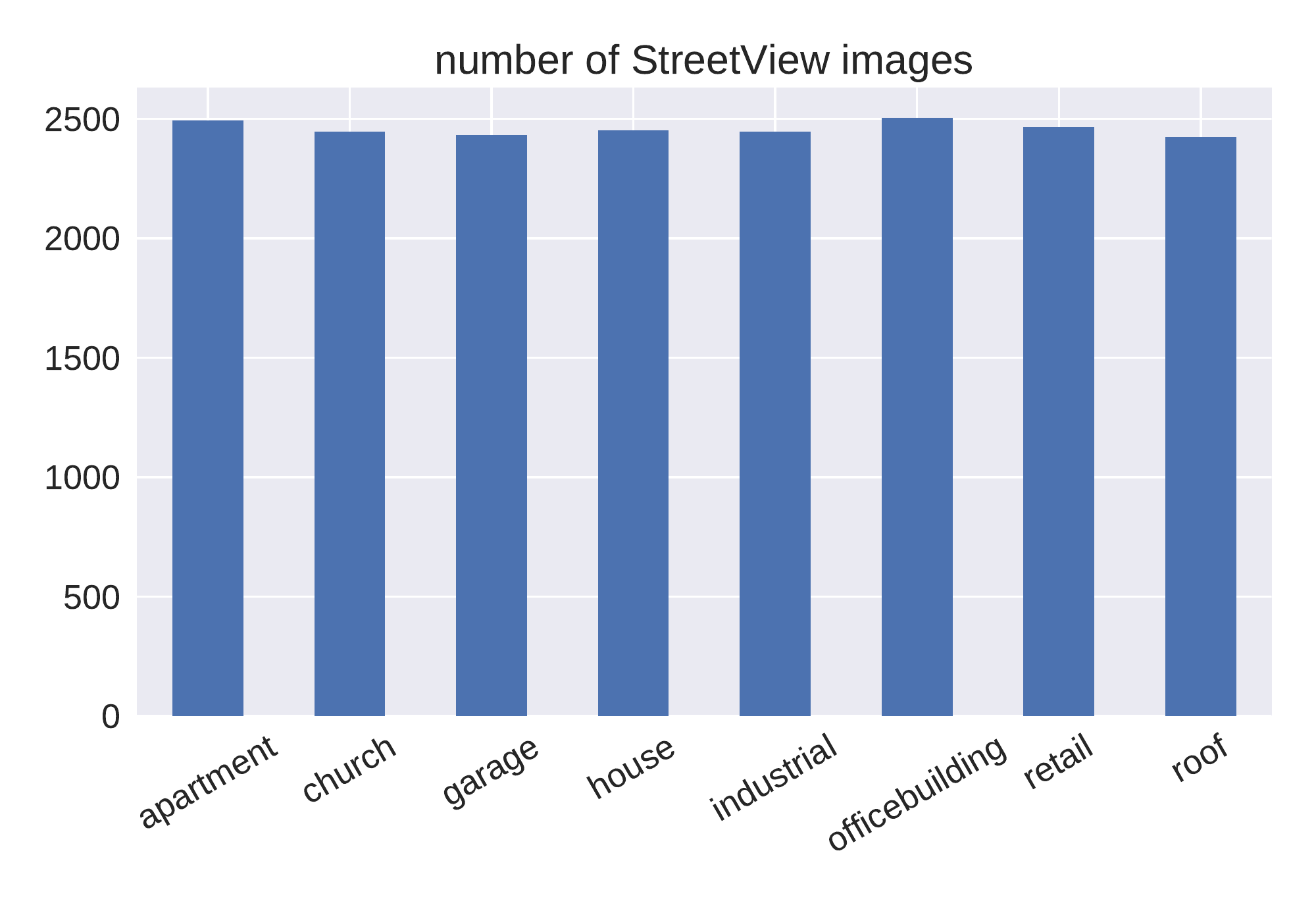}
		\caption{Number of street view images of each building class. }
		\label{fg:each_bd_cls_num}
	\end{figure}

\begin{figure}
		\centering
		\includegraphics[width=0.8\textwidth]{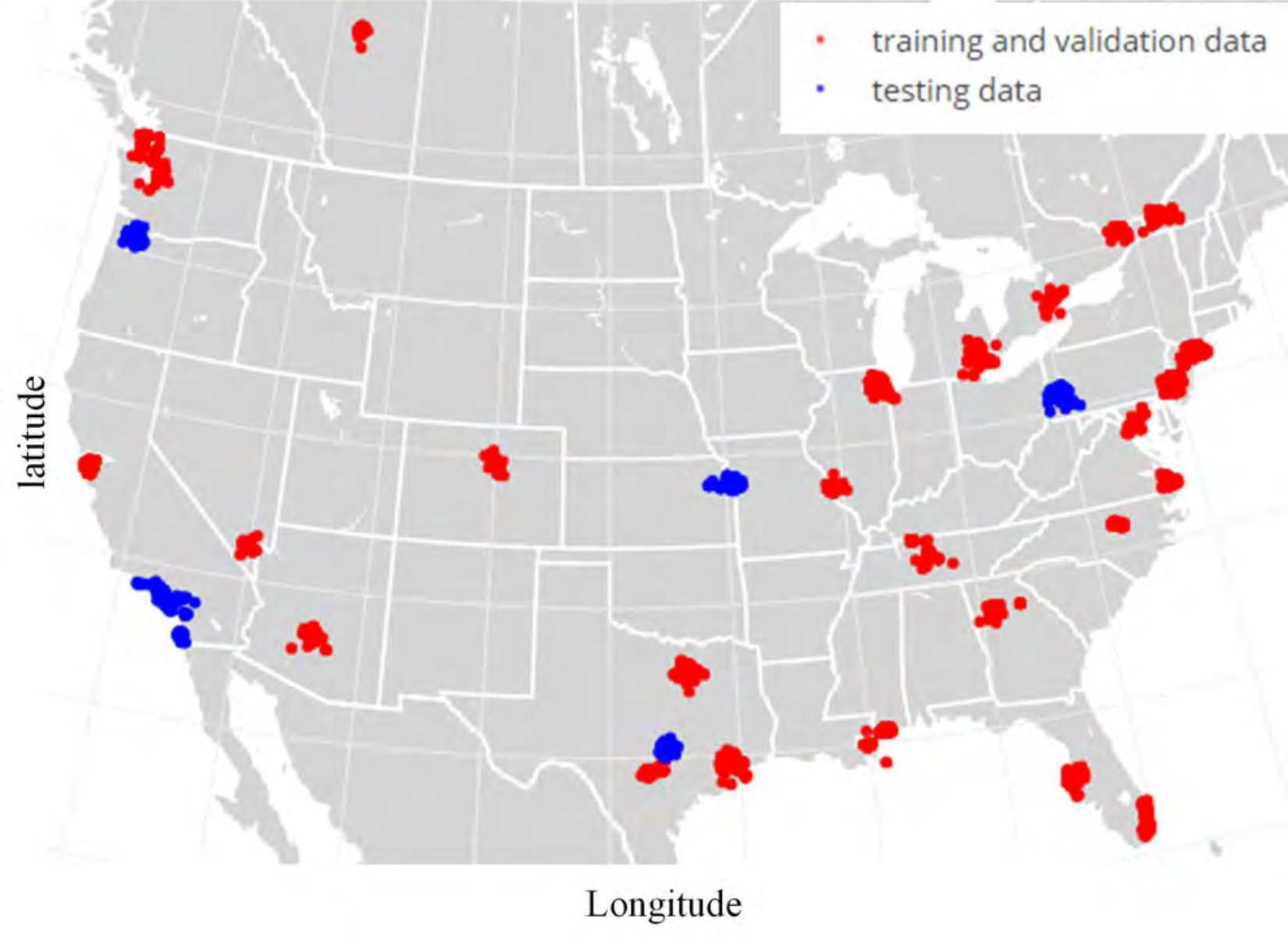}
		\caption{GPS locations of our benchmark dataset. We split all the images into two parts: one for training (17600 images) and the others for testing (2058 images). Note that all the testing images are located in different cities with the training ones.}
		\label{fg:GPS_dataset}
	\end{figure}

	To train a building instance classifier, we first build a corresponding street view benchmark dataset, which contains totally 19658 images from eight classes, i.e. \textit{apartment, church, garage, house, industrial, office building, retail and roof}, and there are around 2500 images for each building class, as shown in Figure \ref{fg:examples_dataset} and \ref{fg:each_bd_cls_num}.  The geo-tagged images are downloaded through Google StreetView API\footnote{\url{https://developers.google.com/maps/documentation/streetview/}}, with the associated metadata\footnote{\url{https://developers.google.com/maps/documentation/streetview/intro}}, i.e. the image size and pitch value are set to be $ 512\times 512$ pixels and $ 10 $ degrees, respectively. As illustrated in Figure \ref{fg:GPS_dataset}, all the street view images are located over several cities of the US and Canada, e.g. Montreal, New York and Denver, and their associated ground truth building labels are extracted from OpenStreetMap\footnote{\url{http://wiki.openstreetmap.org/wiki/Map_Features\#Building}}. The descriptions for the building classes are demonstrated in Table \ref{tb:building_cls_description}.
	
	Since the dataset is not sufficiently large to train a CNN with millions of parameters from the scratch, we choose to fine-tune a pretrained CNN with our dataset. It is common that a pretrained CNN on a large dataset such as \textit{ImageNet} \cite{russakovsky2015imagenet} can be well adapted to other new tasks with small scale datasets, since low-level features such as corners and edges generated by prior layers of CNN are general in different images. The high-level image representations extracted by posterior layers are dependent on different tasks. Therefore, fine-tuning the layers of the pretrained CNN with the new dataset has been proven to be an efficient way for the adaptation of the CNN to a new training task. 
	
\begin{figure}
		\centering
		\subfigure[]{\label{subfig_a}\includegraphics[width=0.44\textwidth]{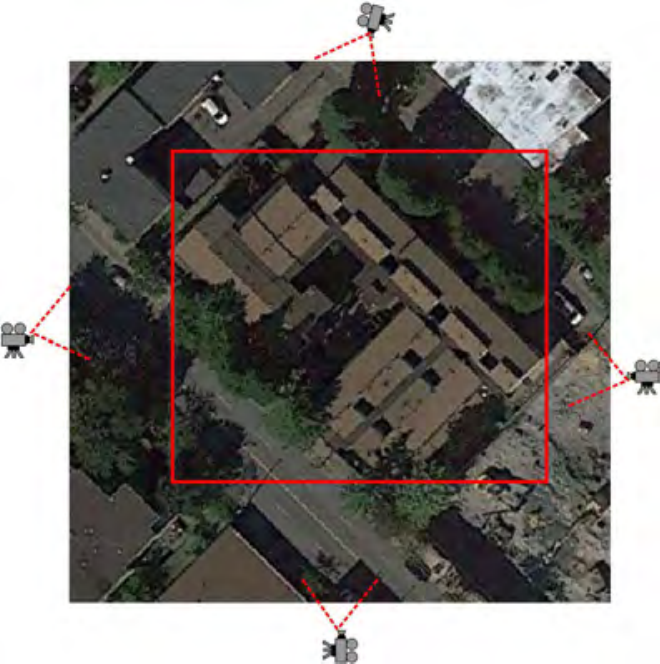}}
		~
		\subfigure[]{\label{subfig_b}\includegraphics[width=0.4\textwidth]{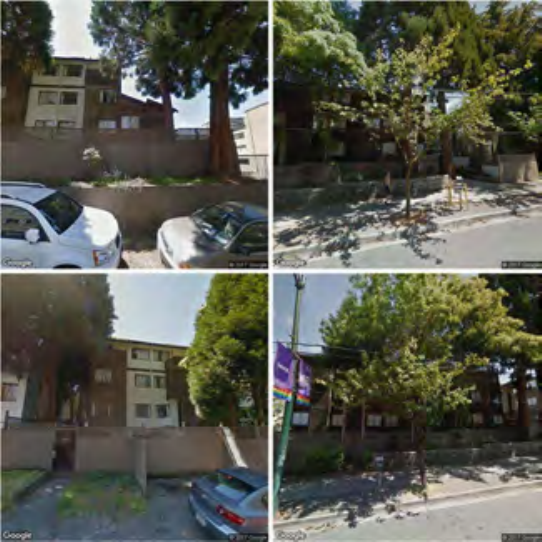}}
		\caption{(a) Illustration of different looking-angles for the same building (red rectangular). (b) The corresponding retrieved street view images: Left column shows the obvious building fa\c{c}ades, while the right two images are outliers. In order to improve the robustness, we classify several street view images for one building, and fuse their classification labels in a decision level.}
		\label{fig_multi_GSV_one_building}
	\end{figure}

	To further improve the classification robustness, the street view images for each building instance are classified, and the building class can be obtained in a decision level. Assuming there are $ M $ street view images retrieved of the study building instance, the final building class $ y $ can be determined by 
		\begin{equation}
		y= \argmax_{i} \frac{1}{M}\sum_{j=0}^{M-1}f^{(j)}_i,
		\end{equation}	
		where $ {f}^{(j)}_i $ is the $ i $th element of the CNN \textit{softmax} layer output $ \mathbf{f}^{(j)} $, which denotes the probability distribution over the whole building classes, and $ j $ is the index of the classified street view image. For example, Figure \ref{fig_multi_GSV_one_building} shows a building to be classified and its corresponding street view images from four different positions. After the right two images filtered by the outlier removal step, we can calculate the final probability distribution vector by averaging those of the left two images and obtain the building class accordingly.

    
	\section{Experiments}
	We train several state-of-the-art CNN architectures, e.g. AlexNet \cite{krizhevsky2012imagenet}, VGG \cite{simonyan2014very} and ResNet \cite{he2016deep} by fine-tuning all the convolutional layers with our benchmark dataset, and demonstrate the corresponding training and testing performances. Among those networks, we choose the best one for generating building classification maps both on region and city scales. 
	\subsection{Training}
	As illustrated in Figure \ref{fg:GPS_dataset}, we split the whole dataset into two parts: 17600 images for training (2200 images for each building class) and 2058 images for testing.  Note that all the testing images are retrieved from different cities with those utilized for training. In order to monitor the training status of networks, we randomly select 3200 images from the training samples to be the validation data. We train four different networks i.e. AlexNet, VGG16, ResNet18 and ResNet34 following the same procedure: Convolutional layers of all these networks are initialized by those pretrained with \textit{ImageNet}, and fully connected layers are randomly initialized following a uniform distribution.

	Each training batch contained in total $ 64 $ images. The stochastic gradient descent algorithm with a learning rate of $ \eta=5\cdot 10^{-4} $ and a momentum value of $ p=0.9 $ was employed for training. To adjust the learning rate, we decayed its value by a factor of $ 0.1 $ in every $ 30 $ epochs. Cross-entropy loss was utilized for training with the weight decay parameter of $ w=10^{-5} $. The neurons of fully connected layers were dropped out by a probability of $ 25\% $. To augment the training data, we randomly cropped $ 224\times224 $ pixels from the original $ 256\times256 $ pixels and randomly flipped the cropped images horizontally. All the experiments were implemented with Pytorch\footnote{\url{http://pytorch.org/}} and carried out by one NVIDIA TITAN X (Pascal) 12GB GPU.
	
\begin{figure}
		\centering
		\includegraphics[width=0.5\textwidth]{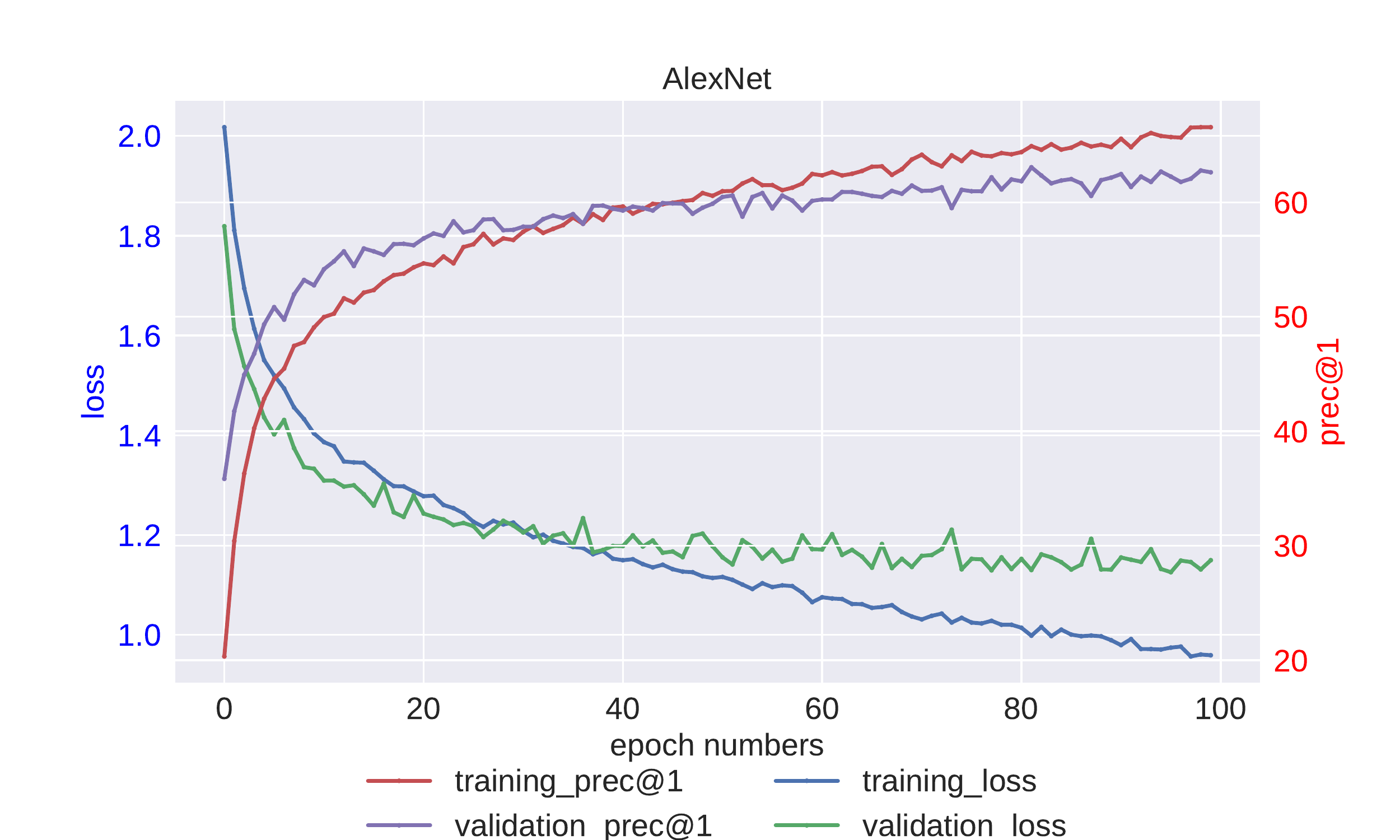}~
		\includegraphics[width=0.5\textwidth]{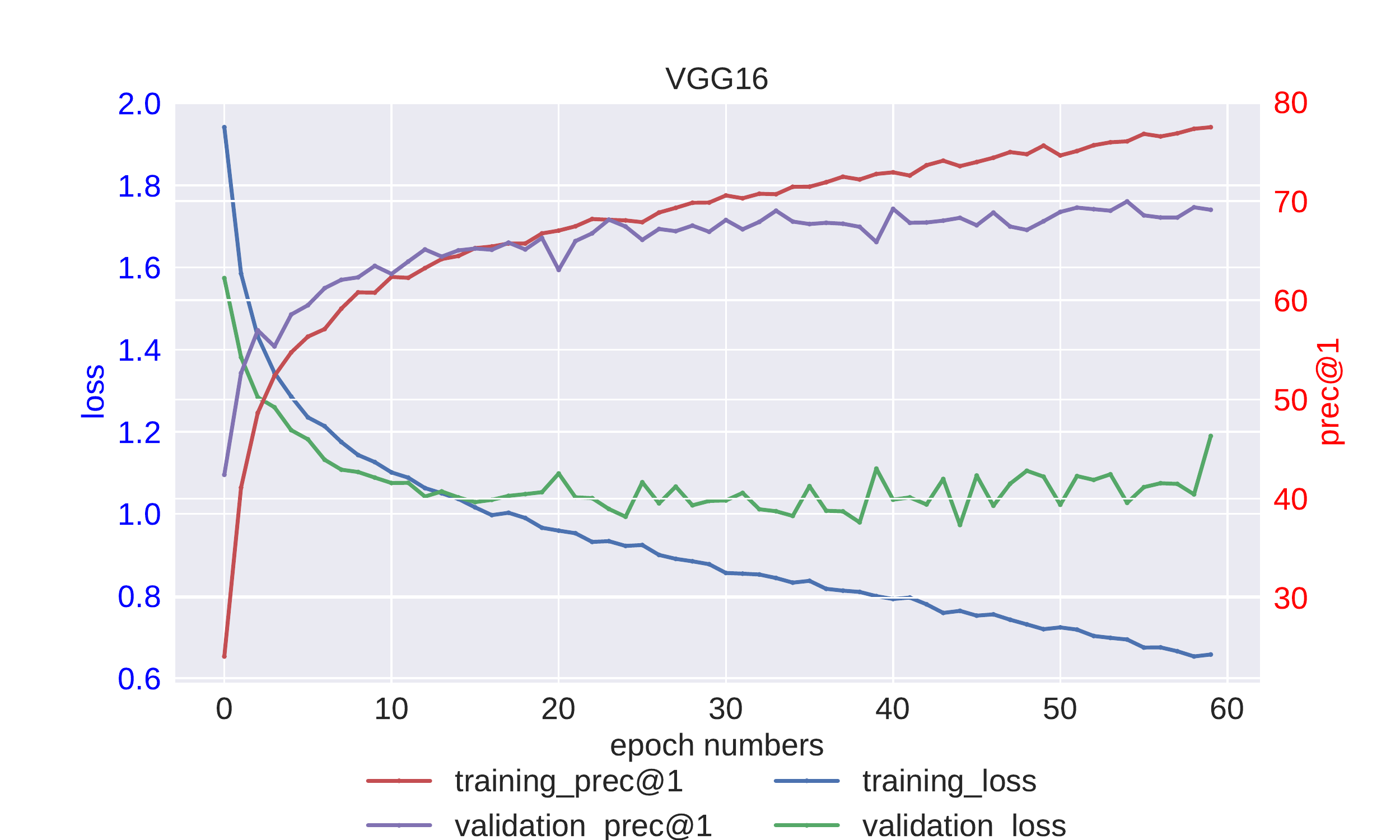}
		
		\includegraphics[width=0.5\textwidth]{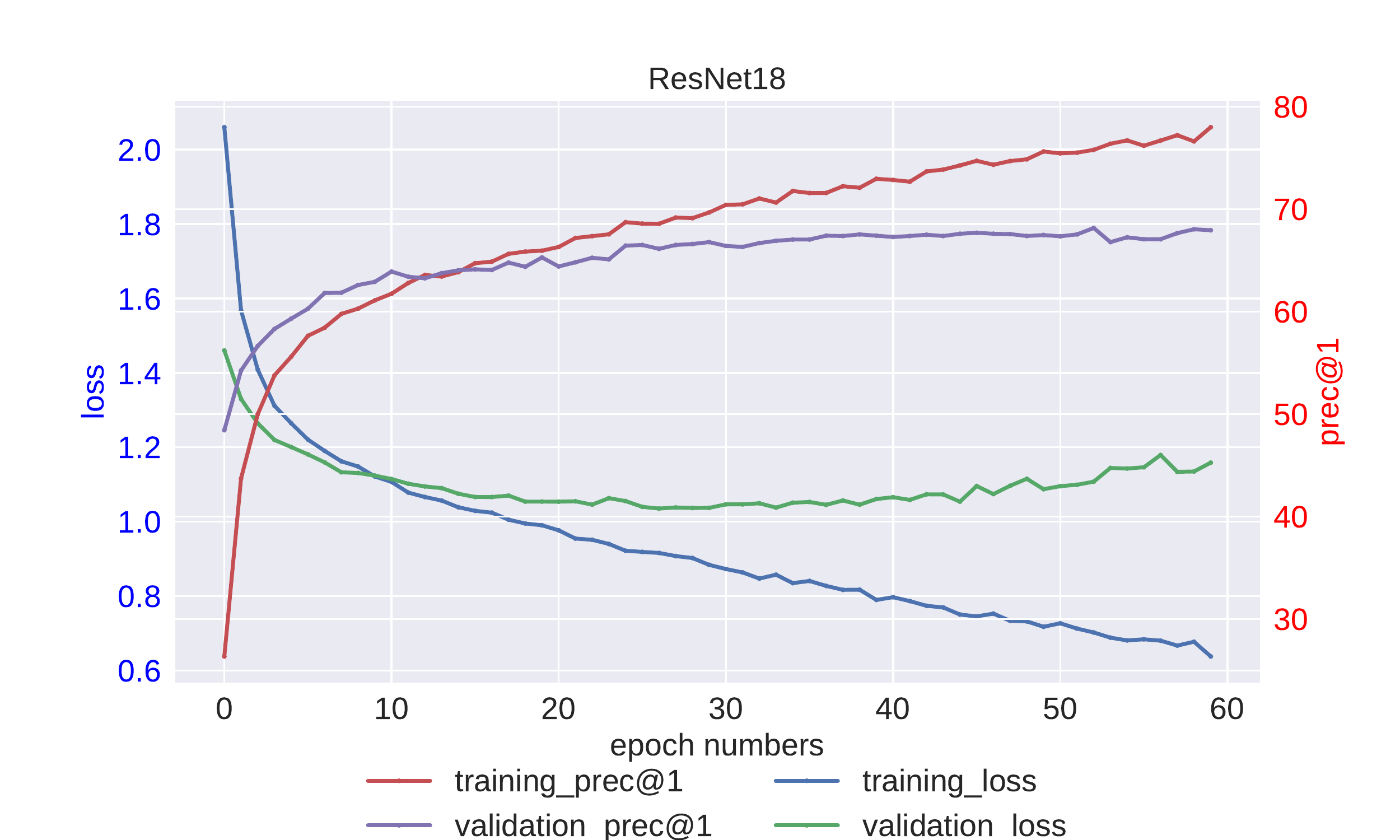}~
		\includegraphics[width=0.5\textwidth]{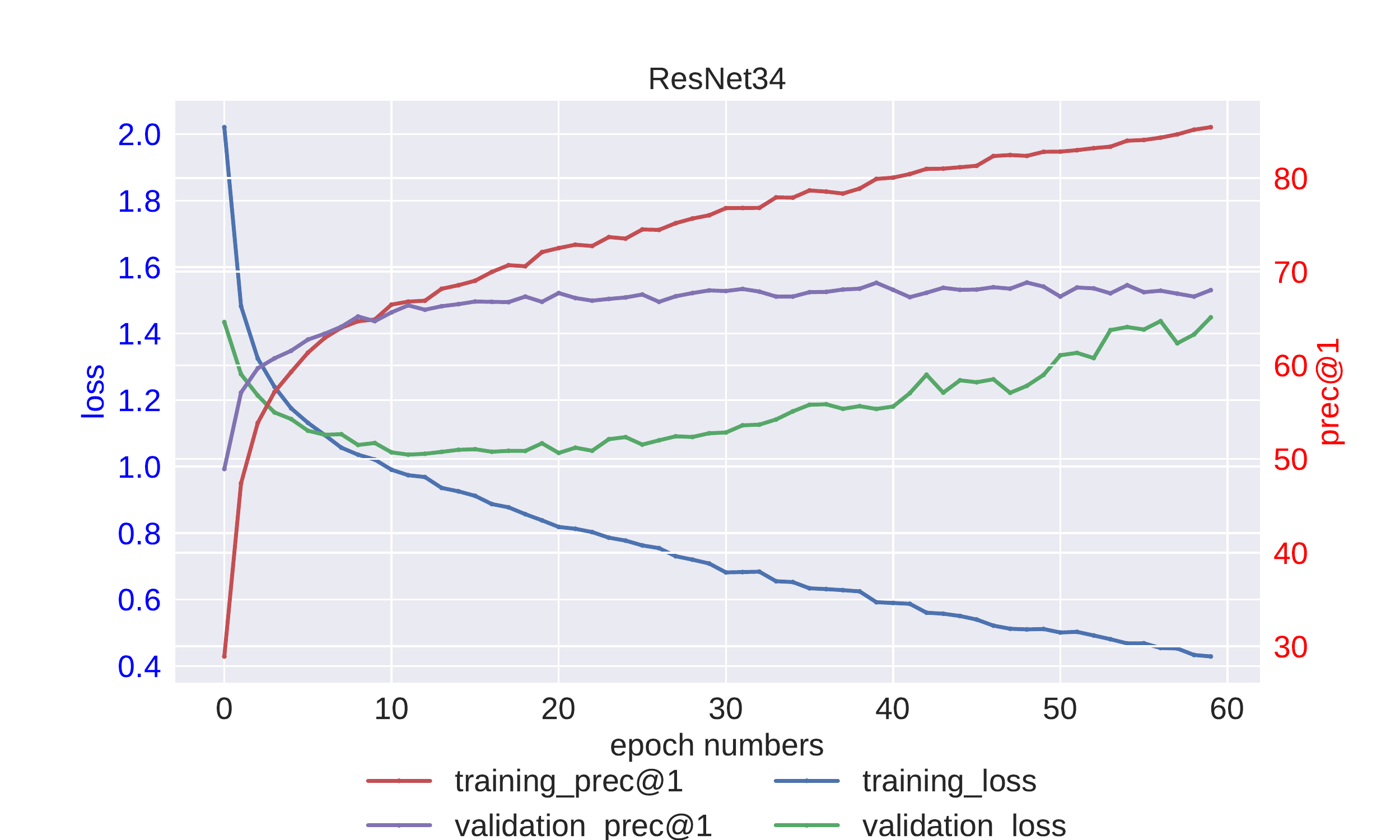}
		\caption{The learning and top 1-precision curves of the four networks, i.e. AlexNet (Top-left), VGG16 (Top-right), ResNet18 (Bottom-left) and ResNet34 (Bottom-right). It can be seen that training losses of the four networks reduce as the epochs increase. Besides, the validation learning curve of AlexNet converges until 80 epochs, and those of the other three networks can converge within 60 epochs. Overfitting behaviors are found in ResNet18 and ResNet34, and it is more severe in ResNet34. One plausible reason is that the total parameter number of ResNet34 (21 million) is more than that of ResNet18 (11 million). As shown by top-1 precisions, AlexNet can achieve about $ 65\% $, while the other networks can obtain about $ 70\% $.}
		\label{fg:learning_curves}
	\end{figure}

	As shown in Figure \ref{fg:learning_curves}, we plot the learning curves of both training and validation data, and calculate the corresponding top 1-precision values during training. It can be seen that training losses of the four networks reduce as the number of epochs increases. Besides, the validation learning curve of AlexNet converges until 80 epochs, and those of the other three networks can converge within 60 epochs. Overfitting behaviors are found in ResNet18 and ResNet34, and it is more severe in ResNet34. One plausible reason is that the total parameter number of ResNet34 (21 million) is more than that of ResNet18 (11 million). As shown by the top-1 precisions, AlexNet can achieve about $ 65\% $, while the other networks can obtain about $ 70\% $. For the follow-up evaluations, we choose ResNet18 trained until 40 epochs and 25 epochs of ResNet34 and compare the performances of those four networks.  
	\subsection{Testing}
\begin{figure}
		\centering
		\includegraphics[width=0.5\textwidth]{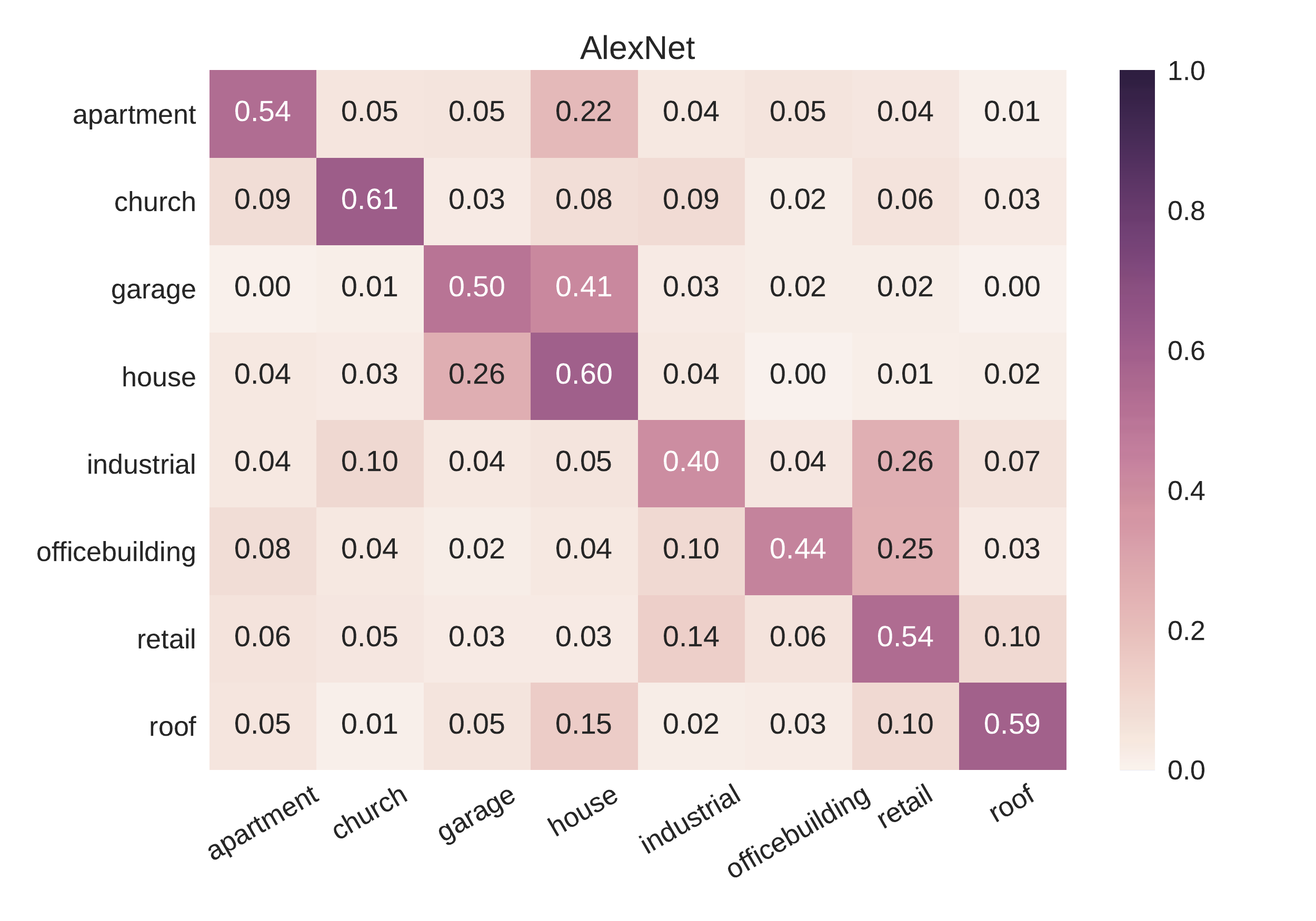}~
		\includegraphics[width=0.5\textwidth]{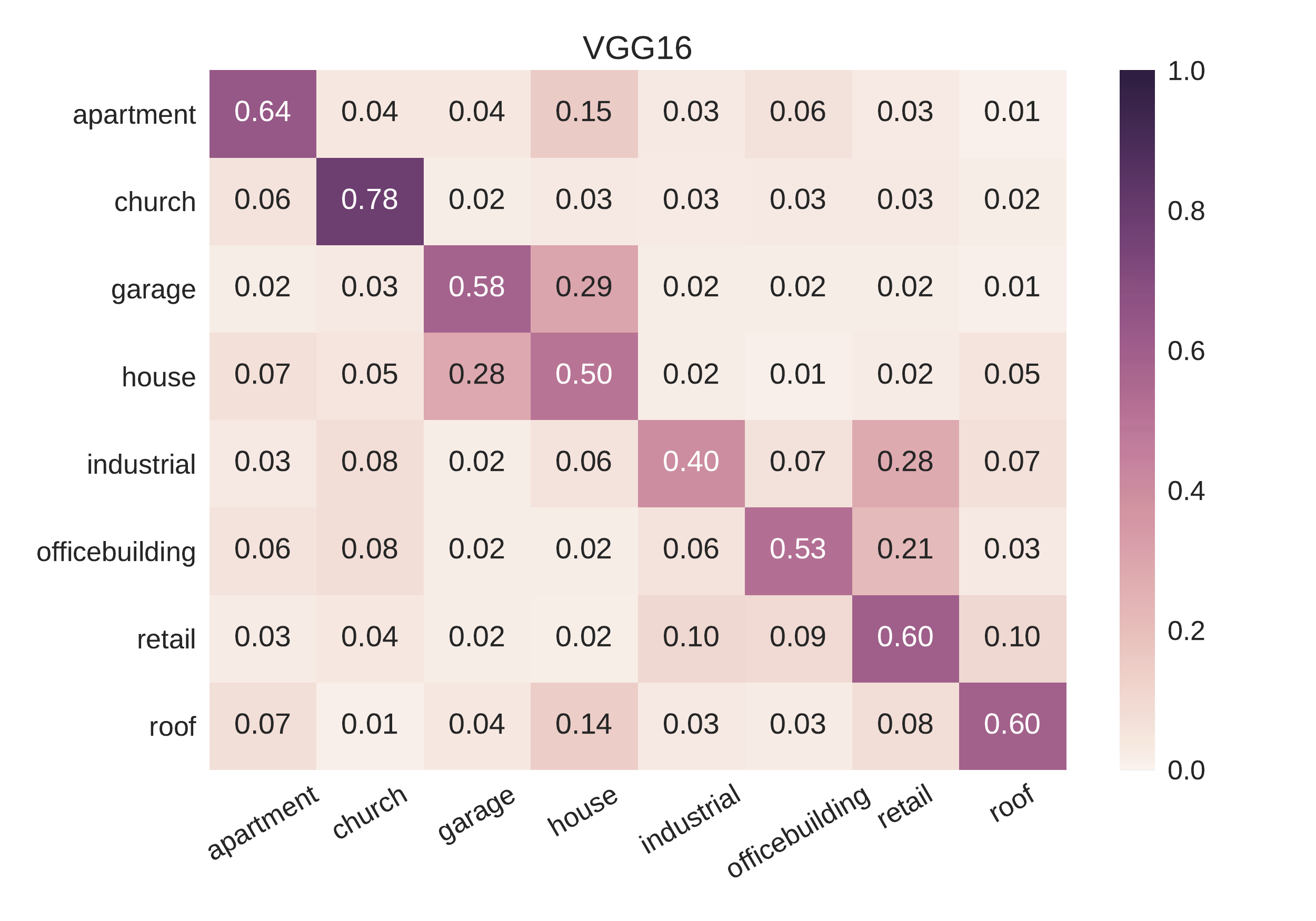}
		
		\includegraphics[width=0.5\textwidth]{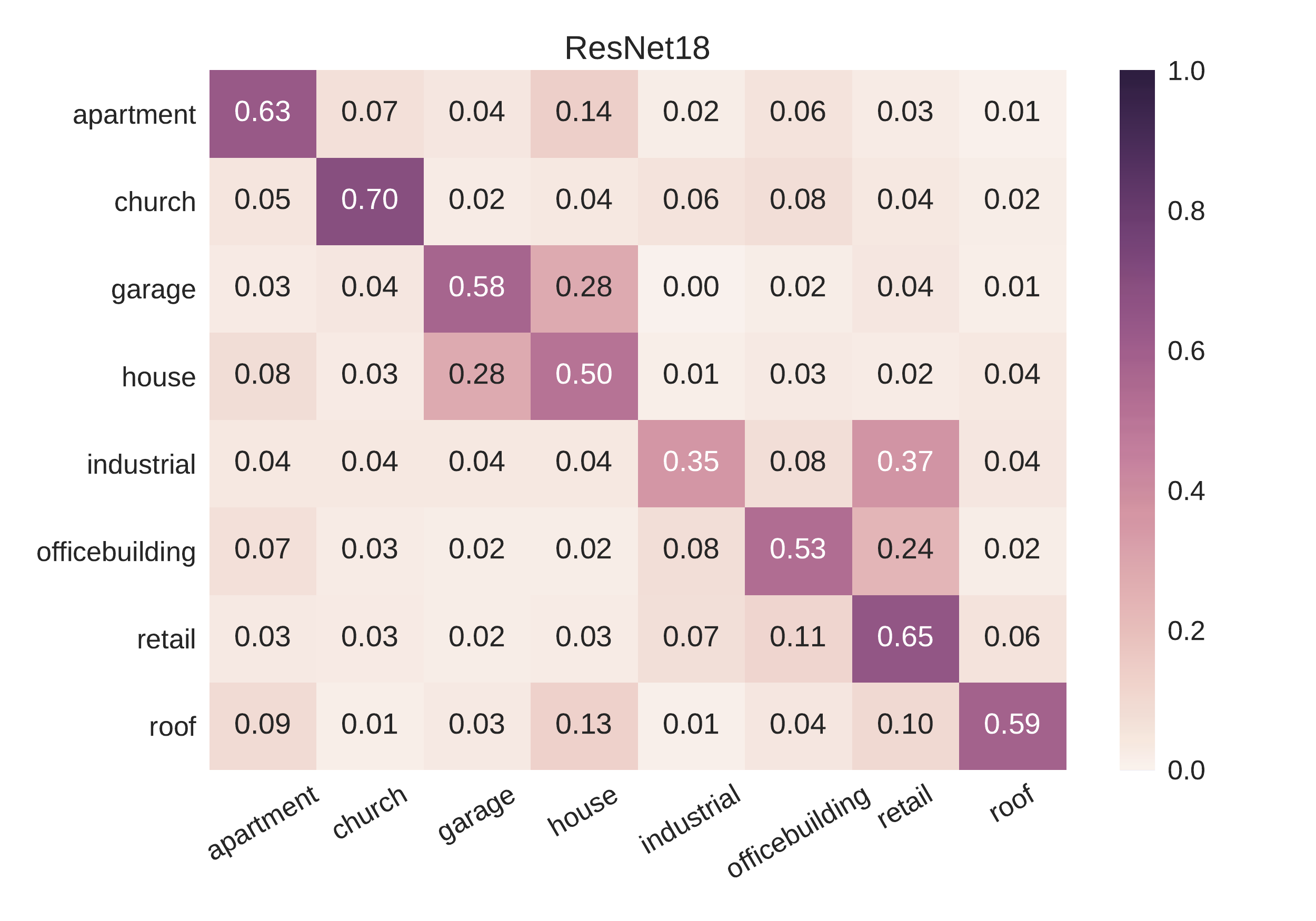}~
		\includegraphics[width=0.5\textwidth]{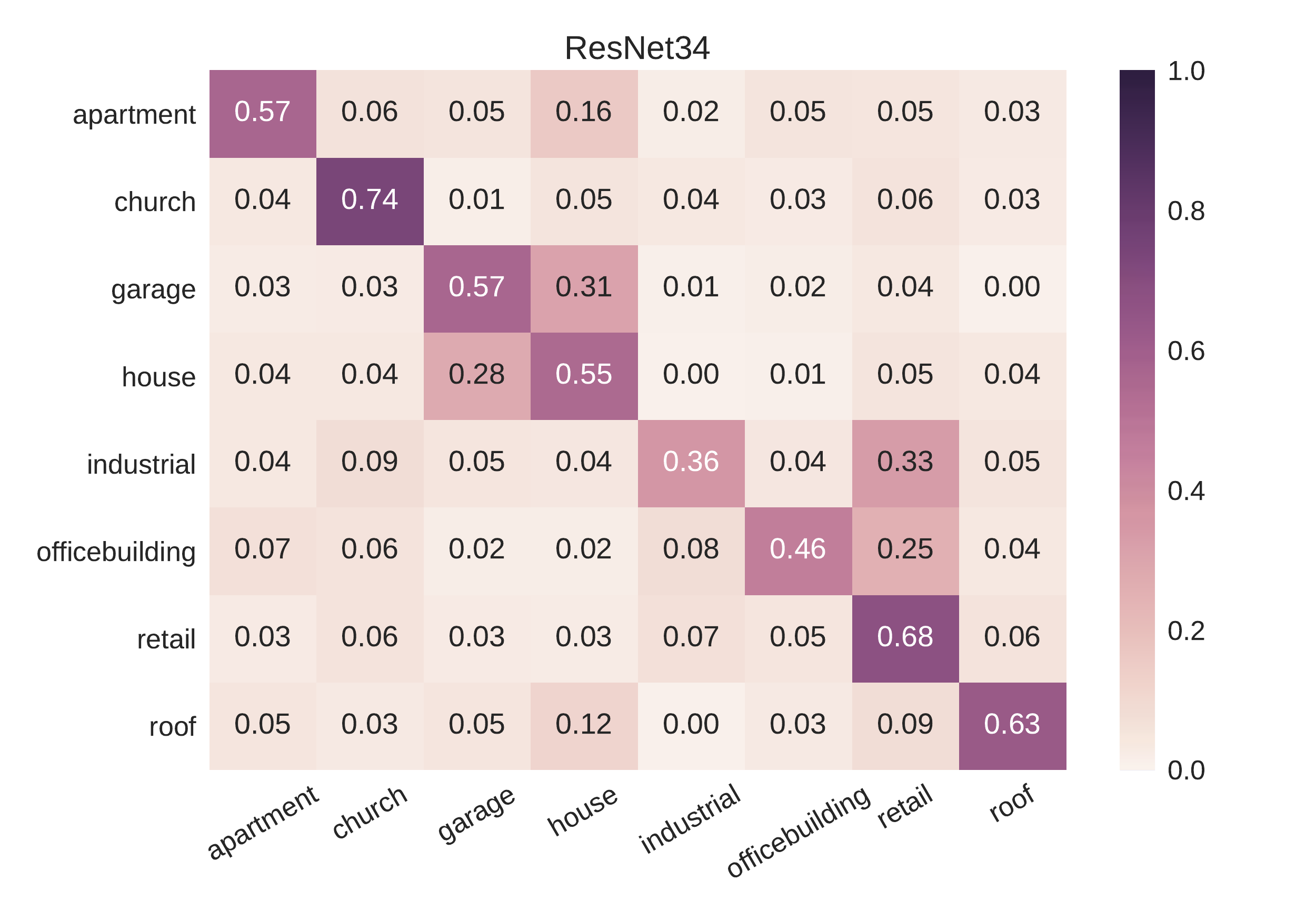}
		\caption{The associated normalized confusion matrices of the four networks evaluated on the test images, i.e. AlexNet (Top-left), VGG16 (Top-right), ResNet18 (Bottom-left) and ResNet34 (Bottom-right).}
		\label{fg:test_img_confusion_mtx}
	\end{figure}

\begin{figure}
		\centering
		\includegraphics[width=0.8\textwidth]{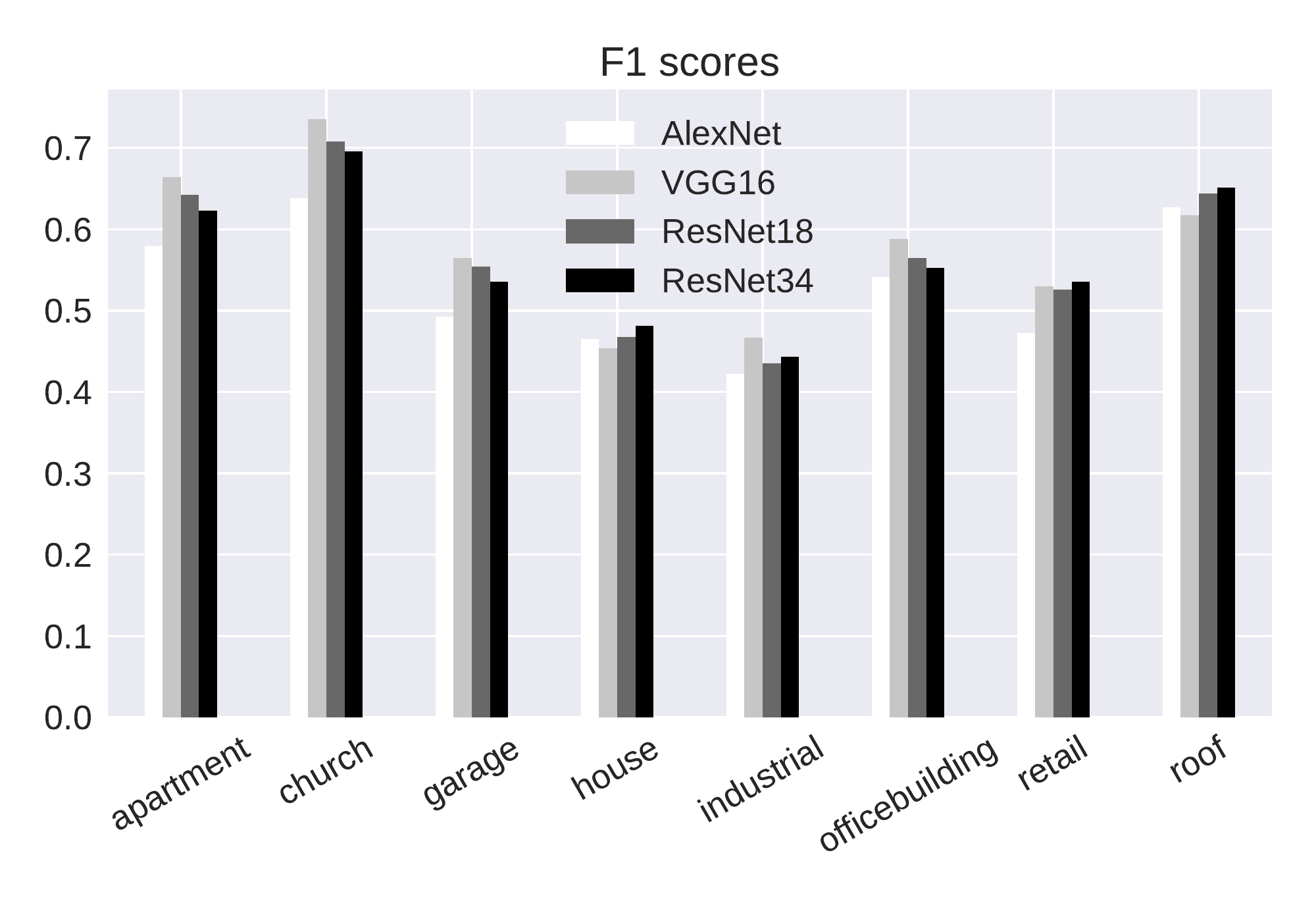}
		\caption{F1 score performances of the four trained networks on the eight building classes. For the classes of apartment, church, garage, industrial and office building, VGG16 achieves the highest F1 score, and for the other classes, ResNet34 is the best among them.}
		\label{fg:f1_score_three_networks}
	\end{figure}

	As illustrated in Figure \ref{fg:test_img_confusion_mtx} and \ref{fg:f1_score_three_networks}, we demonstrate the normalized confusion matrices of all the trained networks evaluated by our test data, and the associated F1 scores of the eight building classes, respectively. F1 score ($ F_1 $), also known as F-measure, is a criteria to measure classification accuracy, which considers both the precision $ p $ and the recall $ r $. It is defined as 
		\begin{equation}
		F_1 = 2\cdot\frac{p\cdot r}{p+r}.
		\end{equation} Moreover, the overall precisions, recalls and F1 scores of the four networks are demonstrated in Table \ref{tb:overall_prec_recall_F1}. From the results, we can see that the classification performance of AlexNet is worse than the other three networks. For the classes of apartment, church, garage, industrial and office building, VGG16 achieves the highest F1 score, and for the other classes, ResNet34 is the best among them. According to the overall accuracies shown in Table \ref{tb:overall_prec_recall_F1}, we choose the trained VGG16 model for the upcoming generation of building classification maps of the study areas.
	
	\begin{table}
		\caption{Overall precisions, recalls and F1 scores}
		\centering
		\begin{tabular}{c|c|c|c}
			\hline
			network & precision & recall & F1 score \\
			\hline
			AlexNet &     $ 0.55 $      &    $ 0.53 $    &     $ 0.53 $     \\
			VGG16   & $ \mathbf{0.59} $ & $ \mathbf{0.58} $ & $ \mathbf{0.58} $ \\
			ResNet18 &     $ 0.58 $     &    $ 0.57 $    &     $ 0.57 $     \\
			ResNet34 &      $ \mathbf{0.59} $    &    $ 0.57 $    &    $ 0.56 $      \\
			\hline
		\end{tabular}
		\label{tb:overall_prec_recall_F1}
	\end{table}
	
	\subsection{Building classification maps of study areas}
	
	\subsubsection{Maps of study areas in Vancouver and Fort Worth}
\begin{figure}
		\centering
		\includegraphics[width=\textwidth]{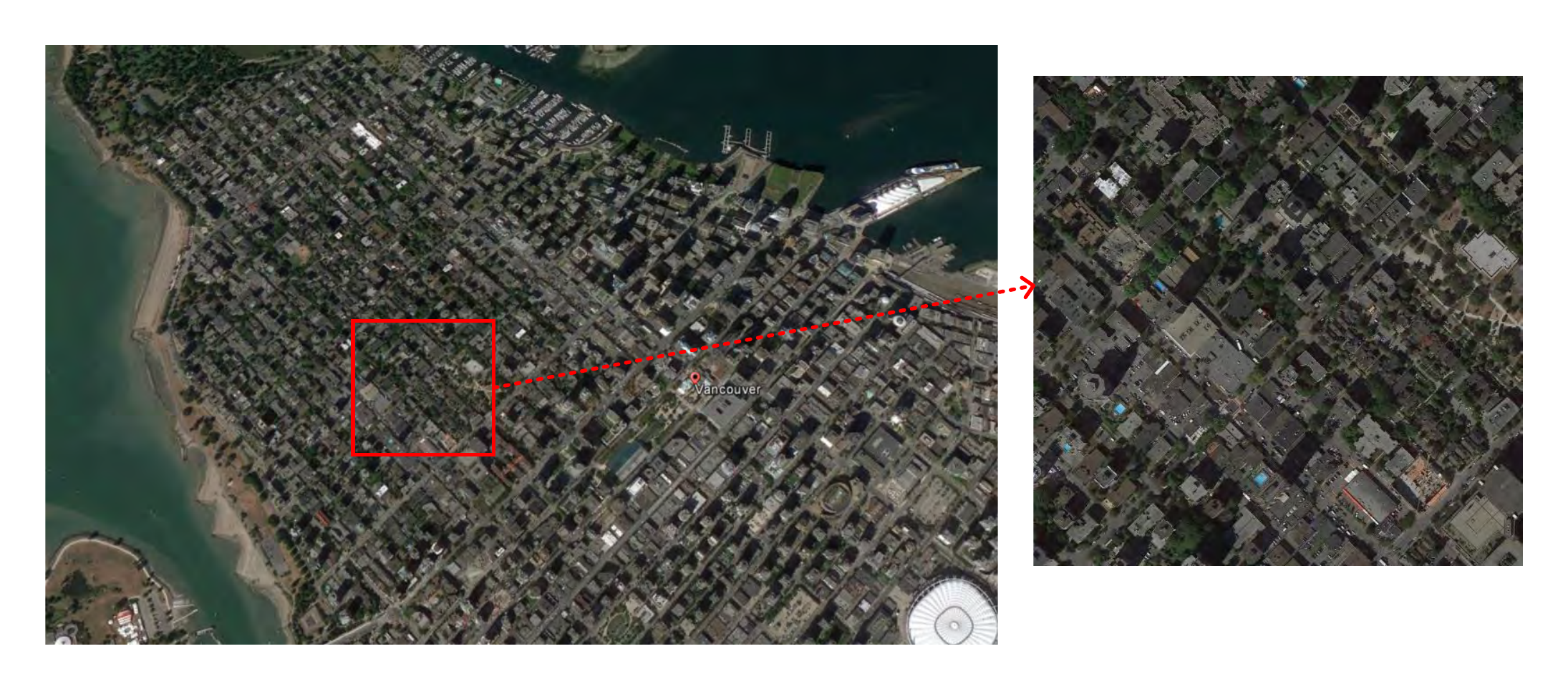}
		\caption{Illustration of one study area in Vancouver (image is from Google Earth).}
		\label{fg:test_area_vancouver}
	\end{figure}

\begin{figure}
		\centering
		\includegraphics[width=\textwidth]{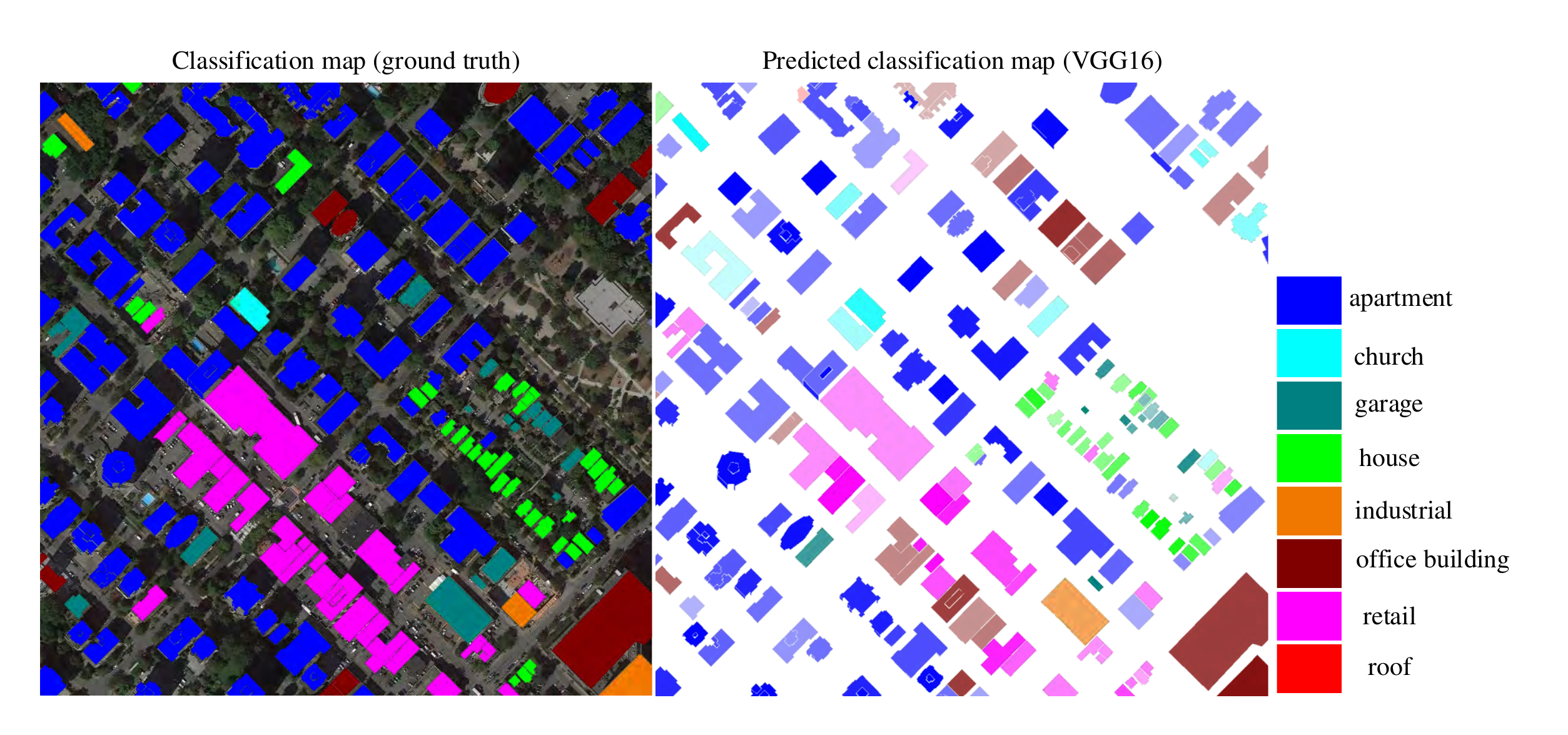}
		\caption{The predicted building classification map (Right), along with the ground truth (Left), where different colors represent different building classes. The total number of building instances in this area is 196. Our result predicts 93 apartments, 10 churches, 13 garages, 24 houses, 1 industrial building, 21 office buildings, 26 retails and 1 roof. 7 buildings are not classified, since no corresponding street view images are found. Moreover, the confidence score for the class of each building is shown by the opacity of the associated color mask, i.e. the higher the opacity, the larger the confidence score and vice versa.}
		\label{fg:GT_predict_vancouver_cls_map}
	\end{figure}

\begin{figure}
		\centering
		\includegraphics[width=0.5\textwidth]{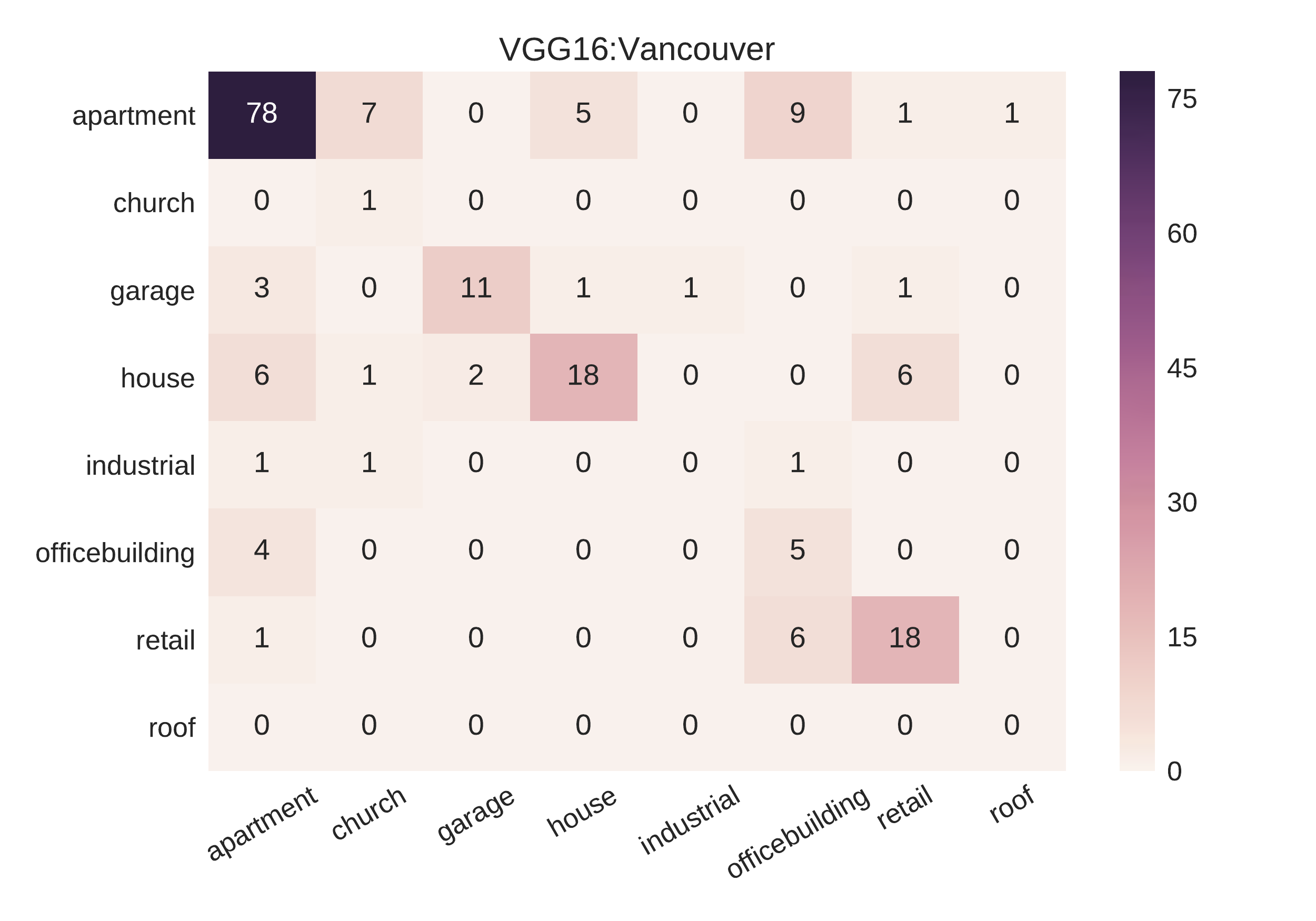}~
		\includegraphics[width=0.5\textwidth]{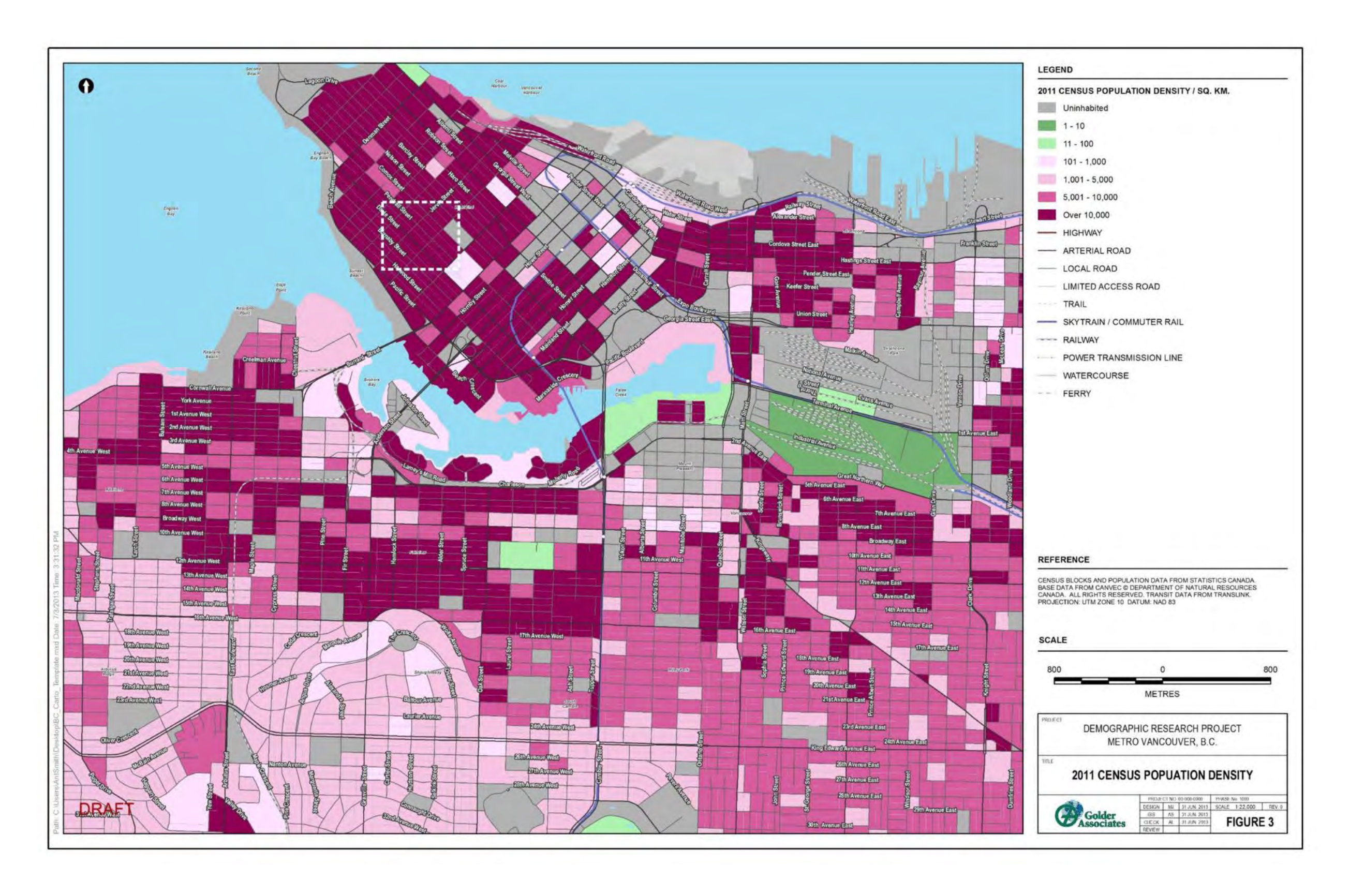}
		\caption{(Left) The confusion matrix of the classification result of the area in Vancouver. We can see that this area is mainly composed by apartments. (Right) 2011 census population density of Vancouver. The white rectangle indicates the study area, which has a high population density of over $ 10000/km^2 $}
		\label{fg:vancouver_confusion_mtx}
	\end{figure}

	One testing area in Vancouver (image is from Google Earth) can be seen in Figure \ref{fg:test_area_vancouver}. The associated ground truth and predicted building classification maps are present in Figure \ref{fg:GT_predict_vancouver_cls_map}, where different colors represent different building classes. We also draw the corresponding confusion matrix of the inferred result in Figure \ref{fg:vancouver_confusion_mtx}. The total number of building instances in this area is 196. Our result predicts 93 apartments, 10 churches, 13 garages, 24 houses, 1 industrial building, 21 office buildings, 26 retails and 1 roof. 7 buildings are not classified, since no corresponding street view images are found. Moreover, the confidence score for the class of each building is shown by the opacity of the associated color mask, i.e. the higher the opacity, the larger the confidence score and vice versa. From the results, we can see that this study area is mainly composed of apartments, which indicates that it is a residential district and there may be a high population density of this area. Our analysis is confirmed by the 2011 census population density of Vancouver downloaded from the website\footnote{\url{https://blogs.ubc.ca/maps/2013/07/03/vancouverpopulationdensity/}}, as shown in Figure \ref{fg:vancouver_confusion_mtx} (Right). The white rectangle in the figure marks the study area which has the highest population density (over $ 10000/\mathrm{km}^2 $). Such classification map gives an insight into the social structure of an residential area. For example, the houses and retails are both grouped together at the right corner of this district. 

\begin{figure}
		\centering
		\includegraphics[width=\textwidth]{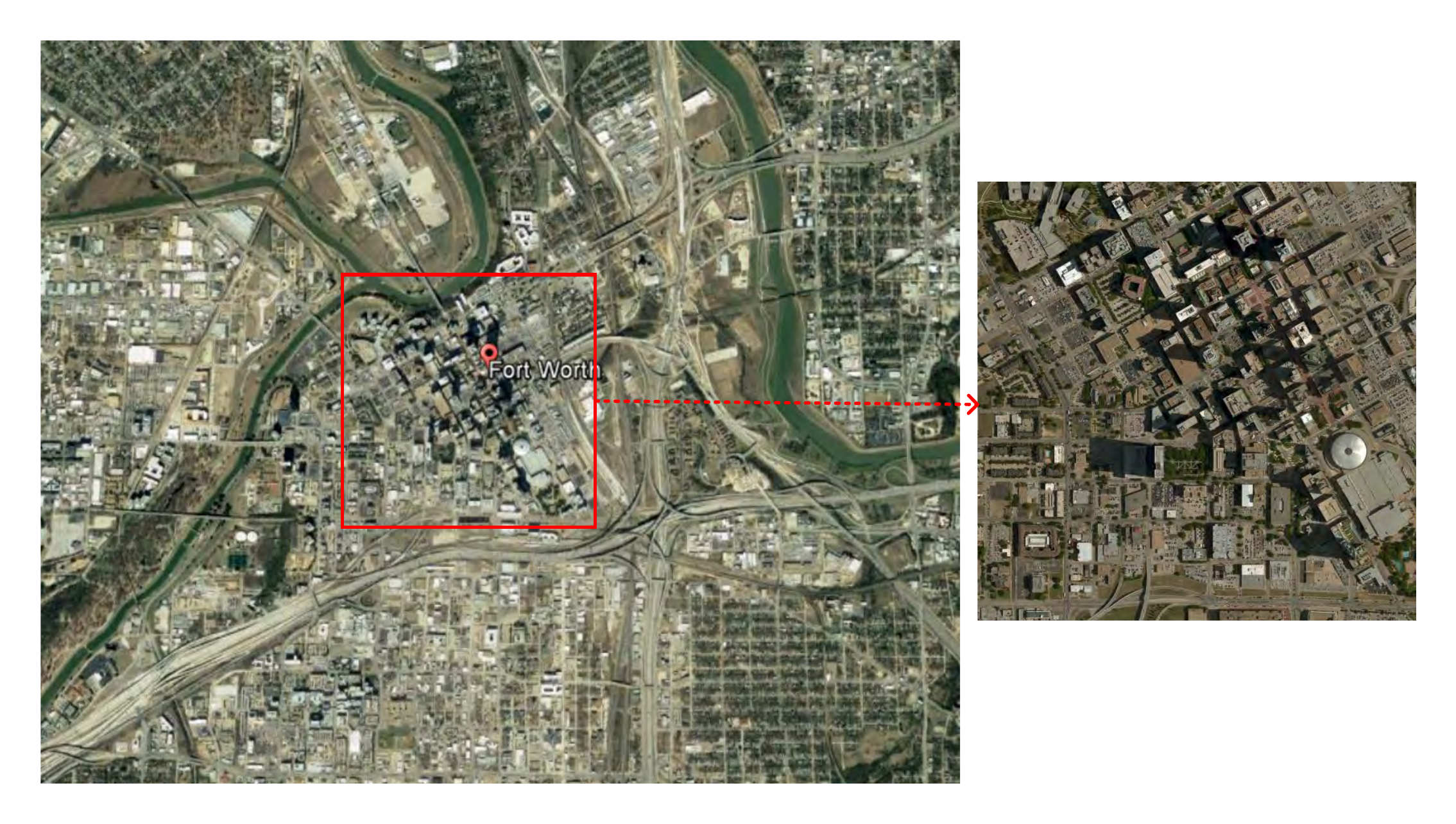}
		\caption{Illustration of one study area in Fort Worth (image is from Google Earth).}
		\label{fg:test_area_fortworth}
	\end{figure}

\begin{figure}
		\centering
		\includegraphics[width=\textwidth]{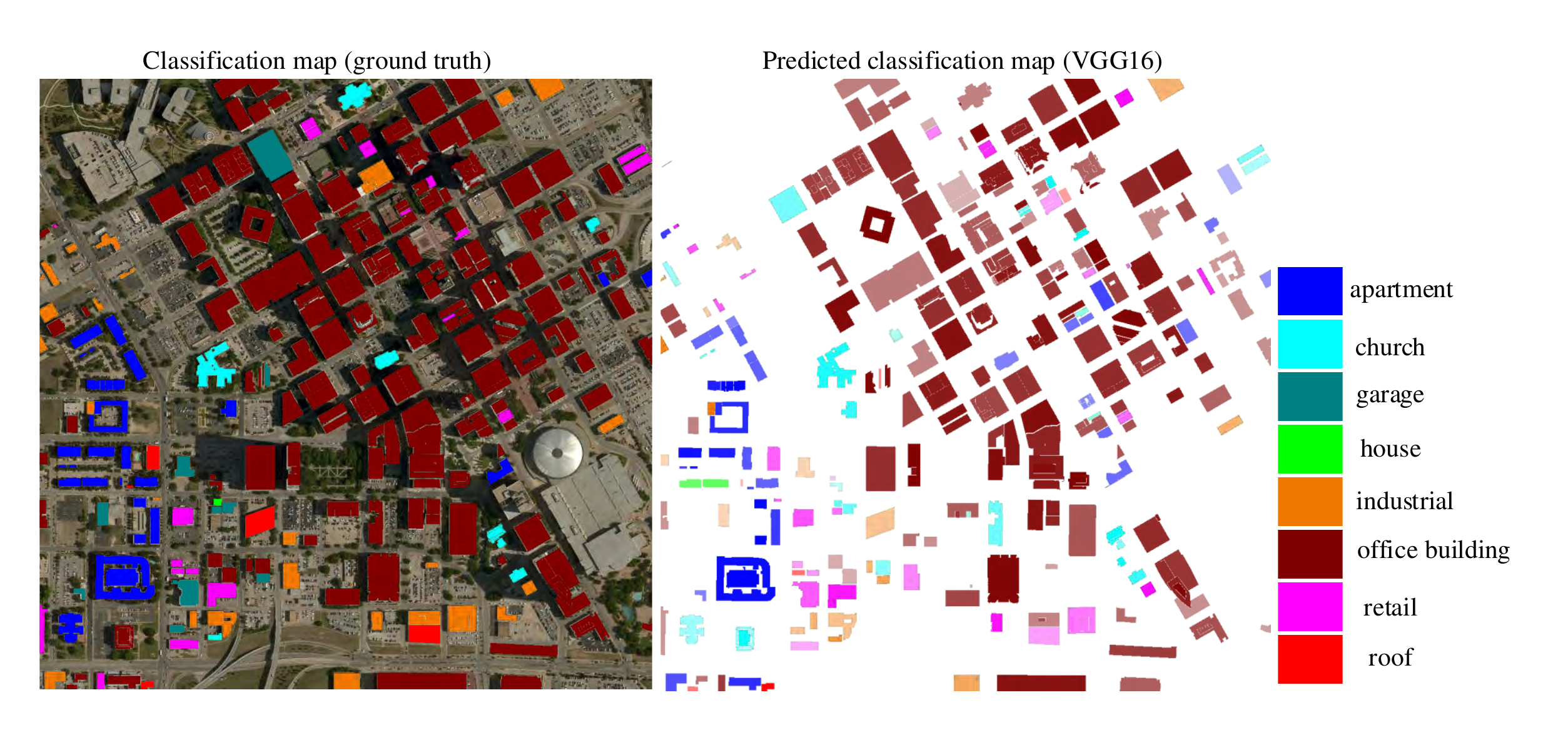}
		\caption{The predicted building classification map (Right), along with the ground truth (Left). The total number of buildings in this area is 316. Our result predicts 34 apartments, 30 churches, 2 houses, 19 industrial buildings, 152 office buildings, 28 retails and 6 roofs. There are no street view images for the remaining 45 buildings.}
		\label{fg:GT_predict_fortworth_cls_map}
	\end{figure}

\begin{figure}
		\centering
		\includegraphics[width=0.6\textwidth]{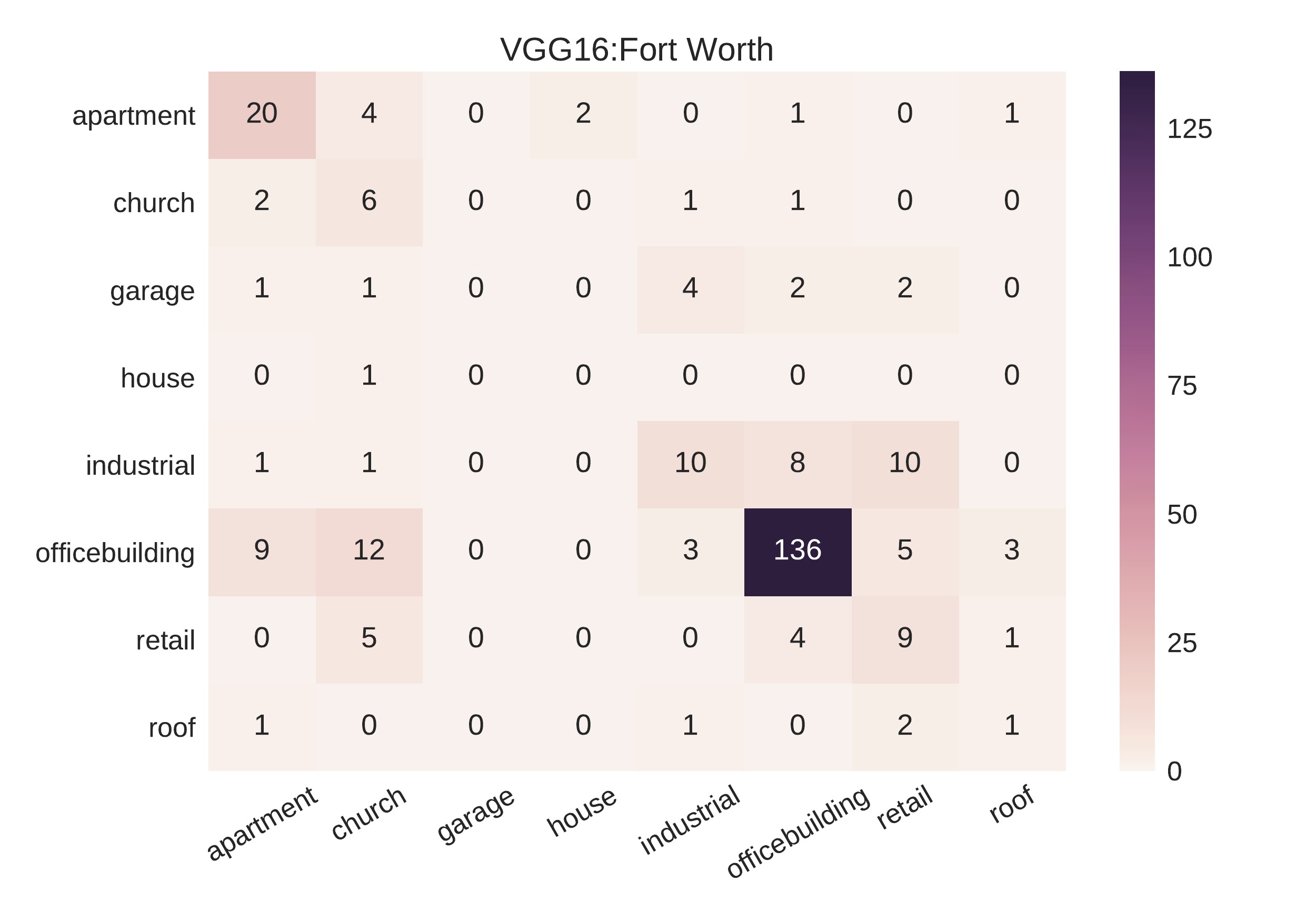}
		\caption{The confusion matrix of the classification result of the area in Fort Worth. We can see that this area is mainly composed by office buildings, which indicates that it is a business district and may locate in the center of Fort Worth.}
		\label{fg:fortworth_confusion_mtx}
	\end{figure}

	Another testing area is located in Fort Worth, shown by the red rectangle area in Figure \ref{fg:test_area_fortworth}. The ground truth and predicted building classification maps are present in Figure \ref{fg:GT_predict_fortworth_cls_map}. The associated confusion matrix is demonstrated in Figure \ref{fg:fortworth_confusion_mtx}. The total number of buildings in this area is 316. Our result predicts 34 apartments, 30 churches, 2 houses, 19 industrial buildings, 152 office buildings, 28 retails and 6 roofs. There are no street view images for the remaining 45 buildings. According to the predicted result, we can see that this area is mainly composed of office buildings, which indicates that it is a business district and may locate in the center of Fort Worth.     
	\subsubsection{City-scale Maps of Calgary, Boston and Toronto}
\begin{figure}
		\centering
		\includegraphics[width=\textwidth]{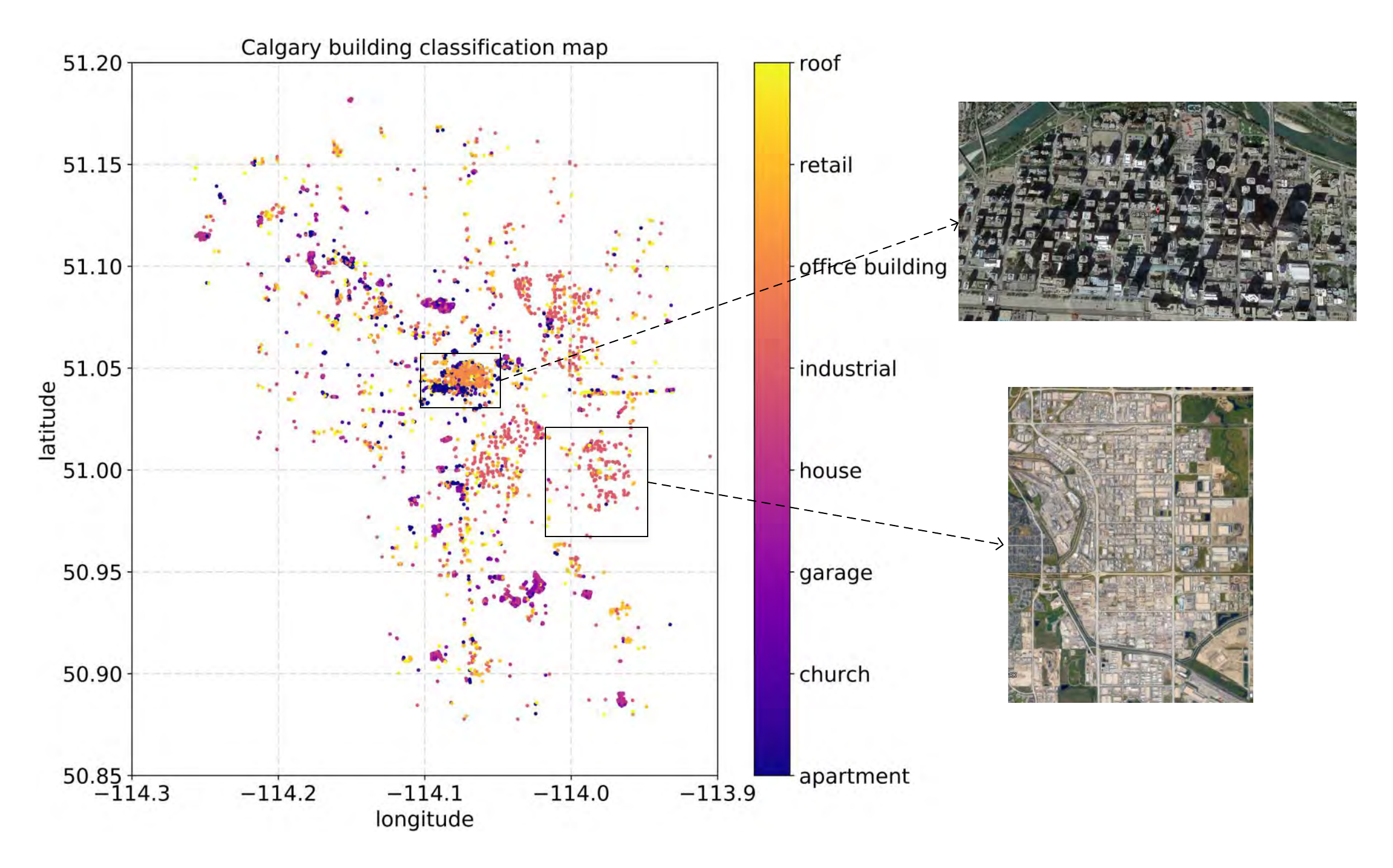}
		\caption{The city-scale building classification map of Calgary, where each classified building instance is displayed as a colored point with GPS coordinates. It is obvious that there are three main industrial districts, and the downtown area is crowded by office buildings. Correspondingly, we also present the remote sensing images of one industrial and the downtown areas (black rectangles). Such classification map can infer that Calgary is an industry city with single central business district to which the three main industrial blocks are located.}
		\label{fg:city_scale_calgary_cls_map}
	\end{figure}

\begin{figure}
		\centering
		\includegraphics[width=0.7\textwidth]{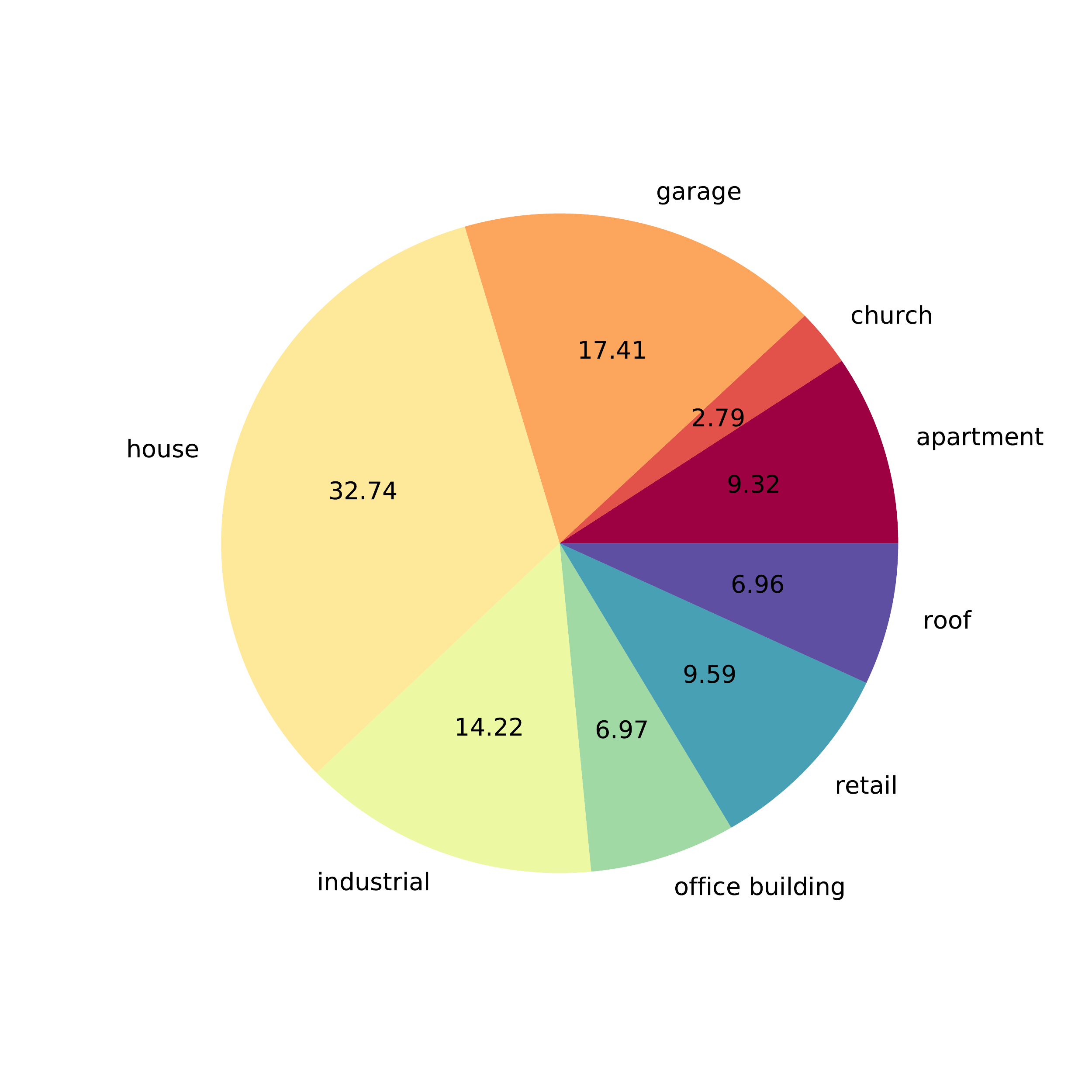}
		\caption{Pie chart of the building class proportions of the predicted buildings of Calgary.}
		\label{fg:calgary_pie_chart}
	\end{figure}

\begin{figure}
		\centering
		\includegraphics[width=\textwidth]{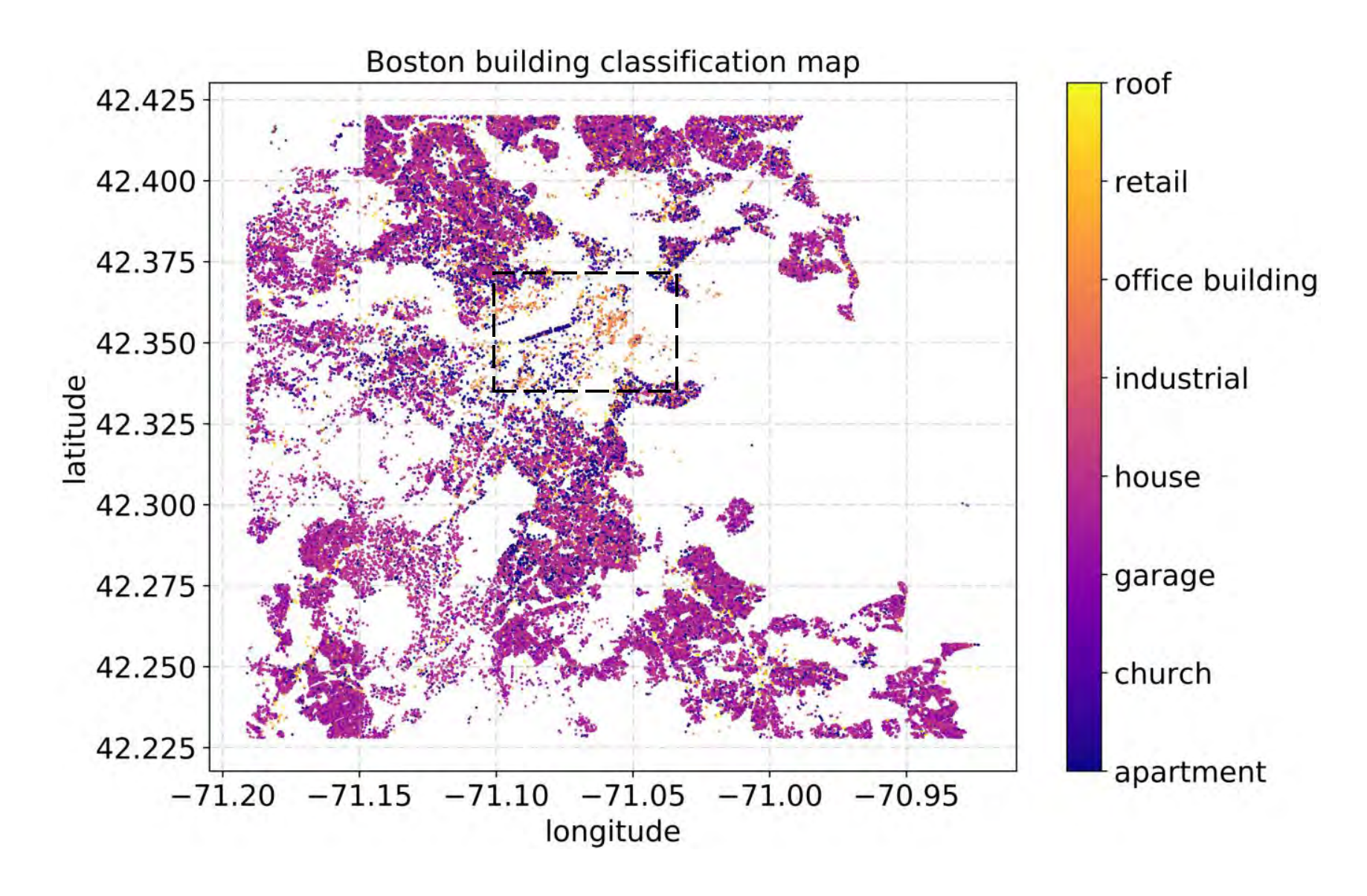}
		~
		\includegraphics[width=\textwidth]{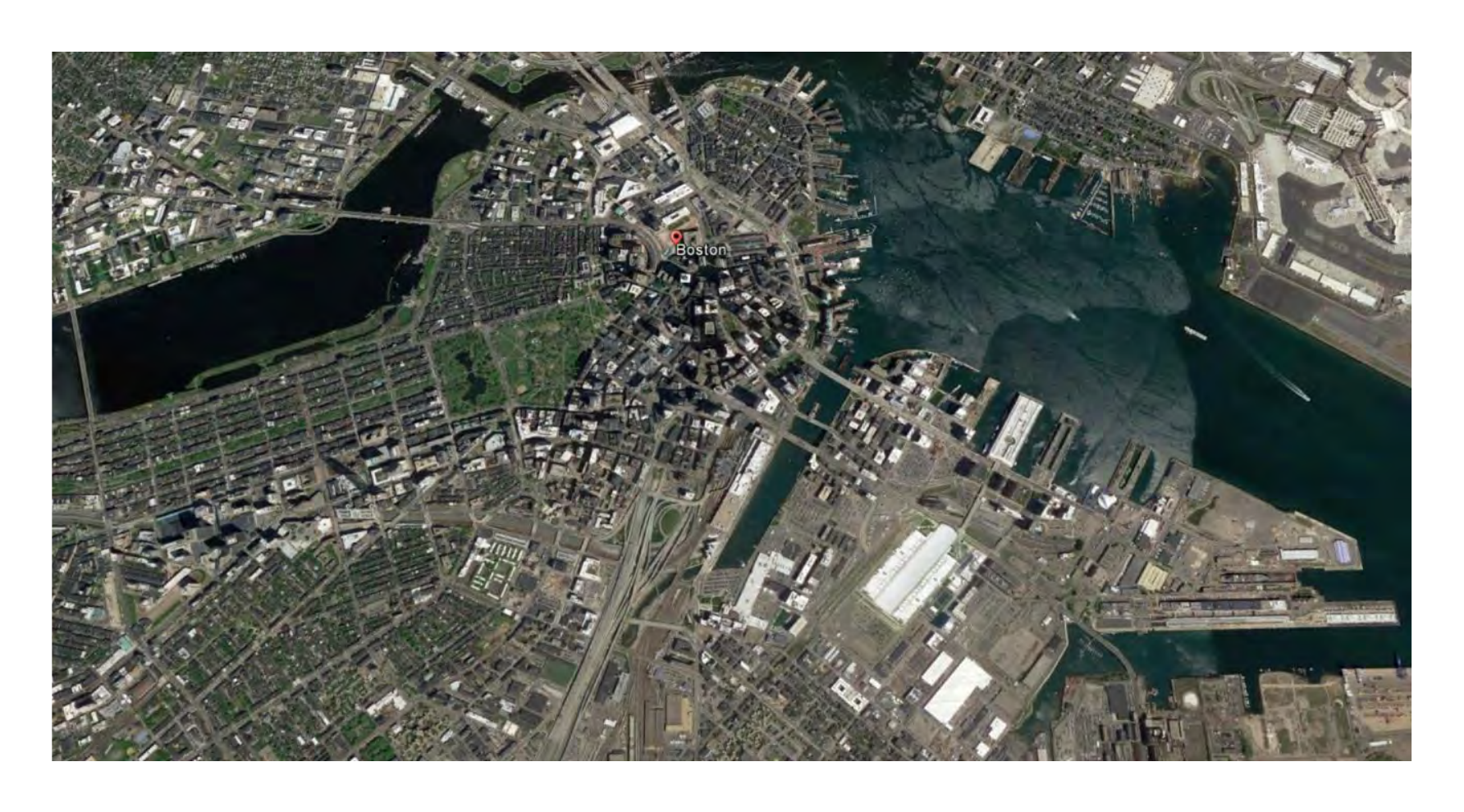}
		\caption{The city-scale building classification map of Boston, where each classified building instance is displayed as a colored point with GPS coordinates. Since most office buildings and apartments locate in the cropped area, it can be observed that Boston is with one single central business district. It is not an industry city, as no large block of industrial districts is found.}
		\label{fg:city_scale_boston_cls_map}
	\end{figure}

\begin{figure}
		\centering
		\includegraphics[width=0.7\textwidth]{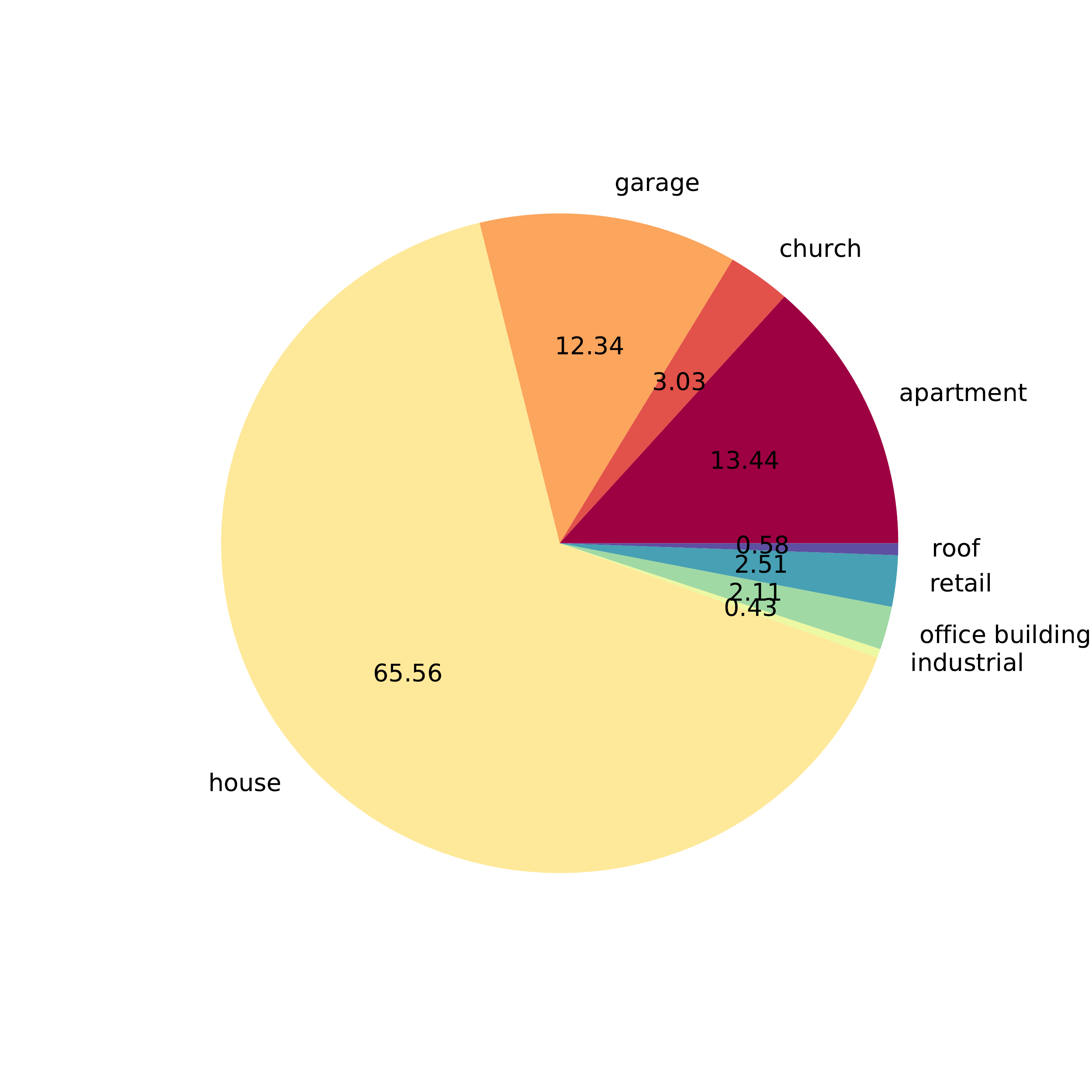}
		\caption{Pie chart of the building class proportions of the predicted buildings of Boston.}
		\label{fg:boston_pie_chart}
	\end{figure}

\begin{figure}
		\centering
		\includegraphics[width=0.8\textwidth]{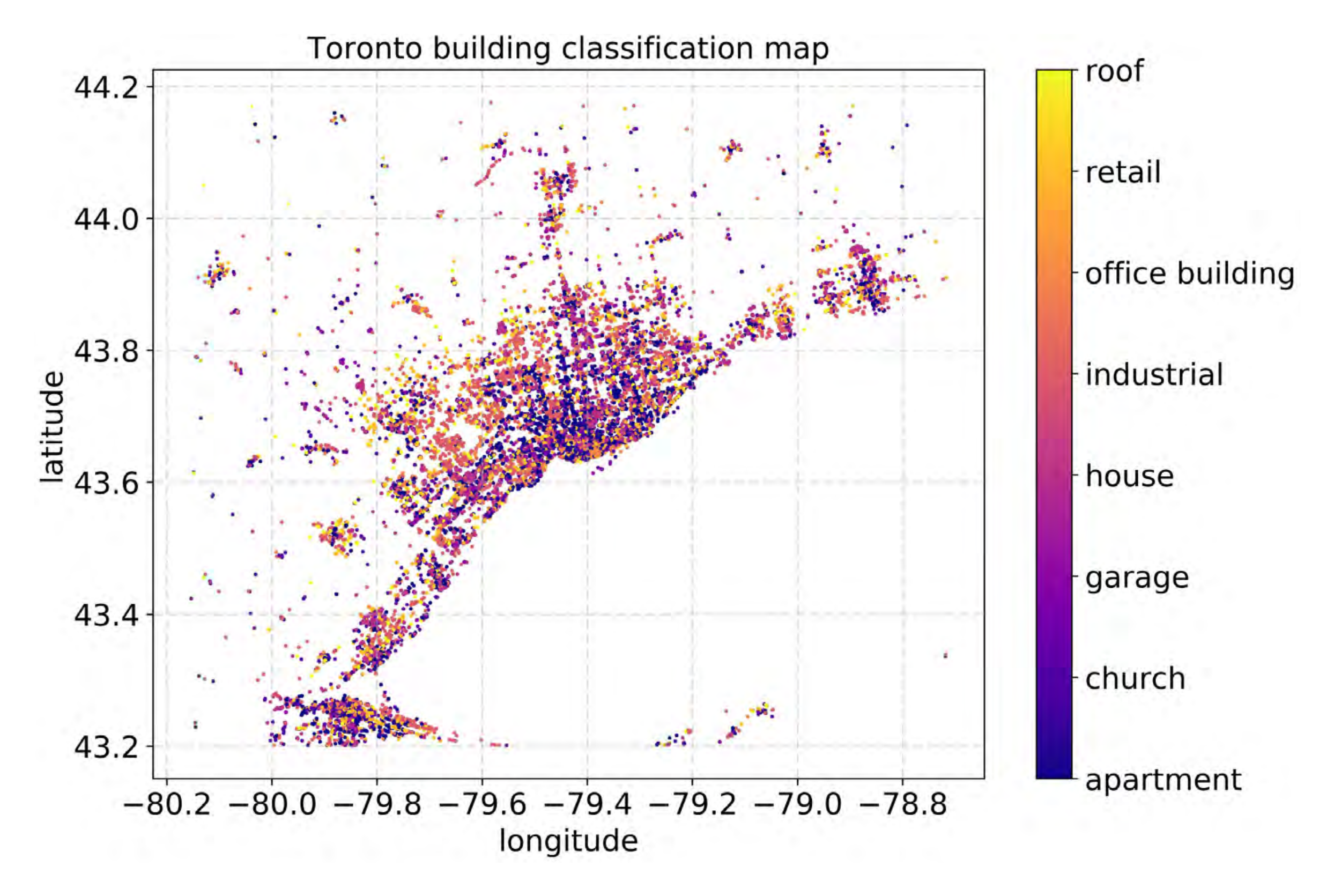}
		~
		\includegraphics[width=0.8\textwidth]{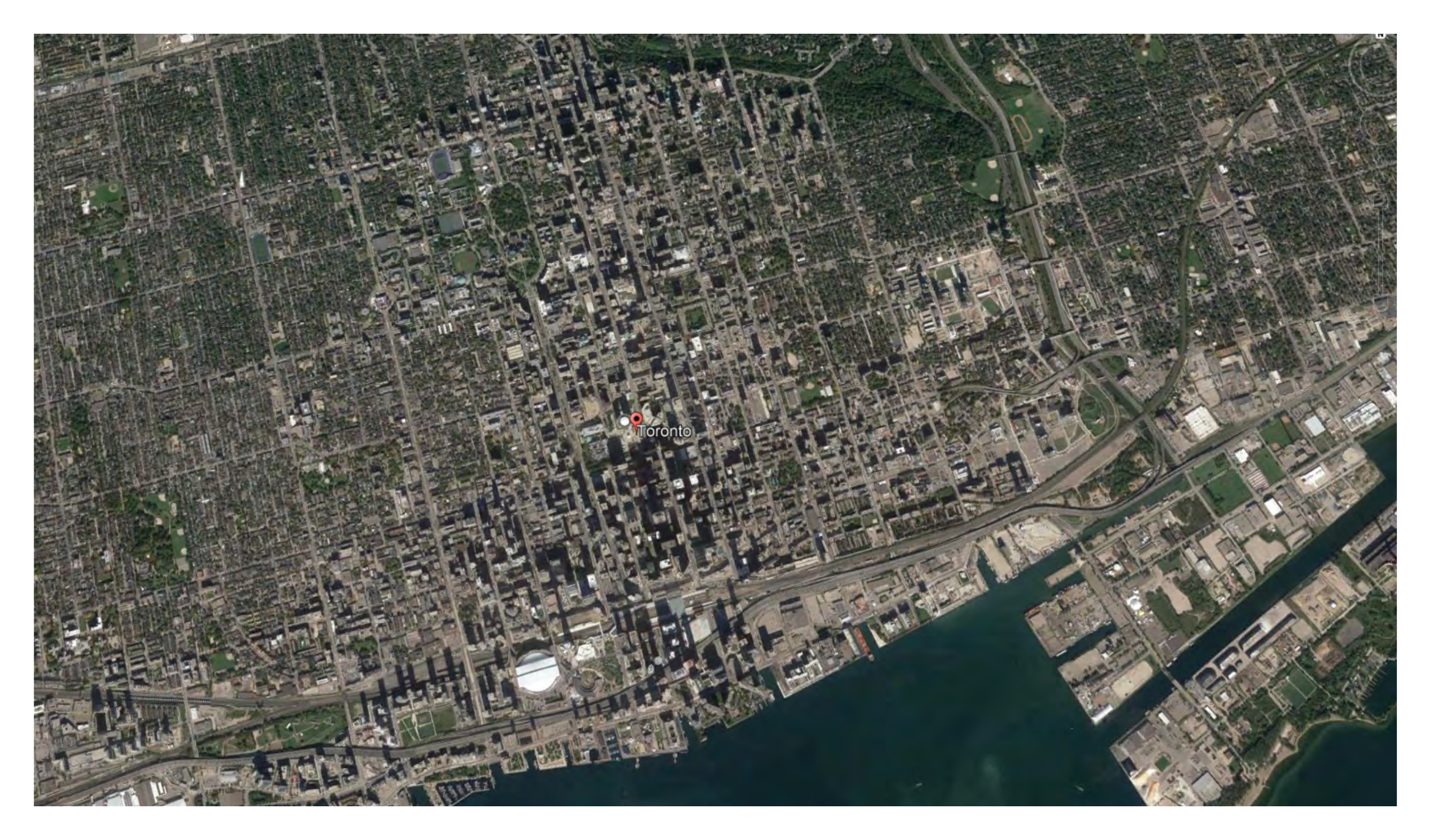}
		\caption{The city-scale building classification map of Toronto and the associated remote sensing image of the central city. Most apartments and office buildings are located in the center of Toronto, and most industrial buildings are distributed in the regions around it.}
		\label{fg:toronto_building_cls_map}
		
	\end{figure}

\begin{figure}
		\centering
		\includegraphics[width=0.7\textwidth]{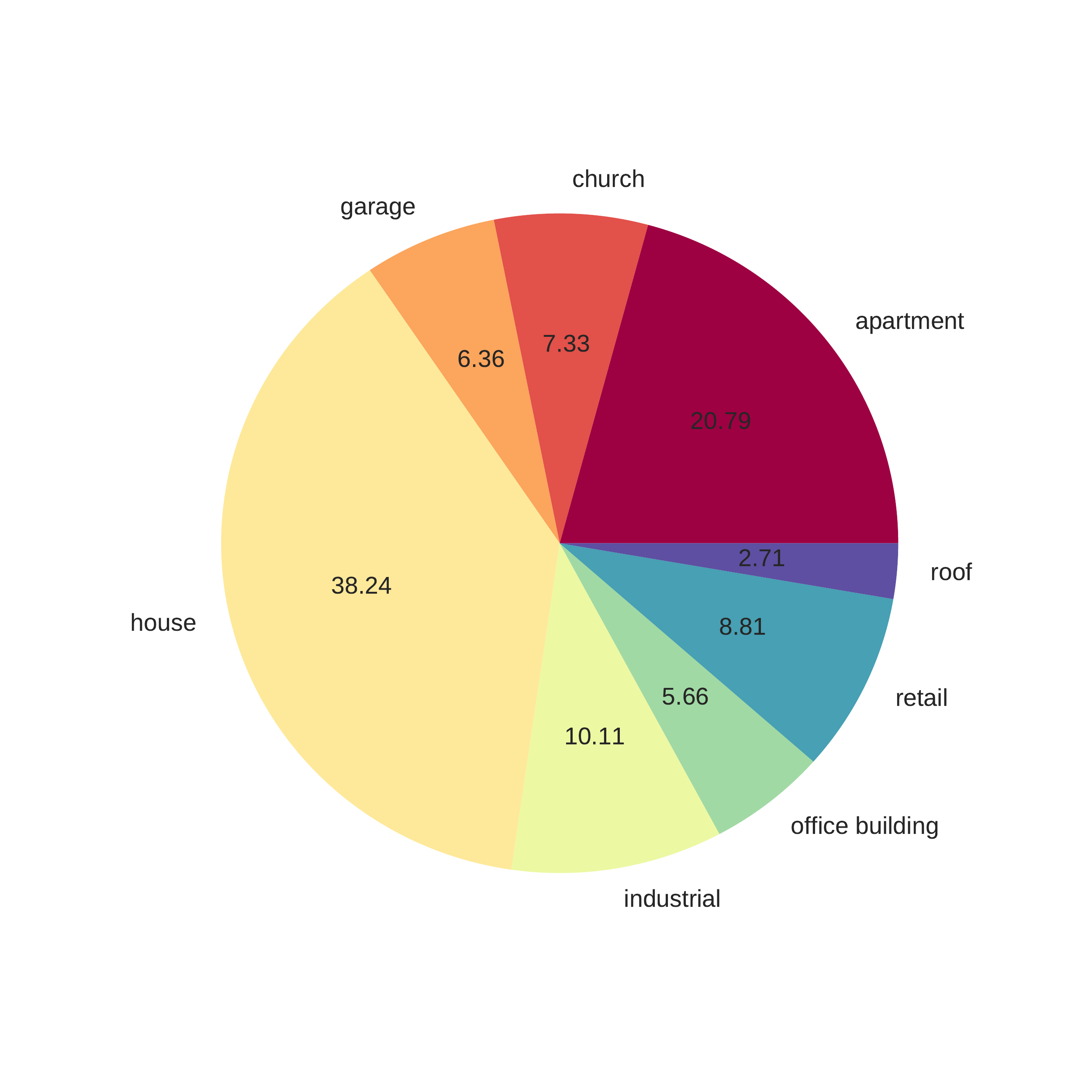}
		\caption{Pie chart of the building class proportions of the predicted buildings of Toronto.}
		\label{fg:toronto_pie_chart}
	\end{figure}

	As shown in Figure \ref{fg:city_scale_calgary_cls_map}, \ref{fg:city_scale_boston_cls_map} and \ref{fg:toronto_building_cls_map}, we provide the city-scale building classification maps of Calgary, Boston and Toronto based on classifying the retrieved 6124, 64389 and 45978 building street view images, respectively, where each classified building instance is displayed as a colored point with its GPS coordinate. Besides, Figure \ref{fg:calgary_pie_chart}, \ref{fg:boston_pie_chart} and \ref{fg:toronto_pie_chart} demonstrate the associated numerical proportions of building classes based on the classification results. In order to quantitatively analyze the performance, 1000 buildings in each city are randomly selected and their associated building tags from OSM are retrieved according to their GPS locations. The classification performances of the three cities are demonstrated in Table \ref{tb:cls_perform_Calgary}, \ref{tb:cls_perform_Boston} and \ref{tb:cls_perform_Toronto}, respectively. Table \ref{tb:cls_perform_Calgary} demonstrates that the overall accuracy of the classification result in Calgary is around $ 0.7 $, given the retrieved $ 1000 $ building tags from OSM. As illustrated in Table \ref{tb:cls_perform_Boston}, by comparing with the building tags from OSM, the overall accuracy of the classification map in Boston can reach around $ 0.55 $. According to Table \ref{tb:cls_perform_Toronto}, more than $ 75\% $ buildings in Toronto can be accurately classified. 
	
	As shown by the classification map of Calgary, it is obvious that there are three main industrial districts, and the downtown area is crowded by office buildings. Correspondingly, we also present the remote sensing images of one industrial and the downtown areas (black rectangles). Such classification map can infer that Calgary is an industry city with single central business district to which the three main industrial blocks are located.
	
	
\begin{figure}
		\centering
		\includegraphics[width=0.3\textwidth]{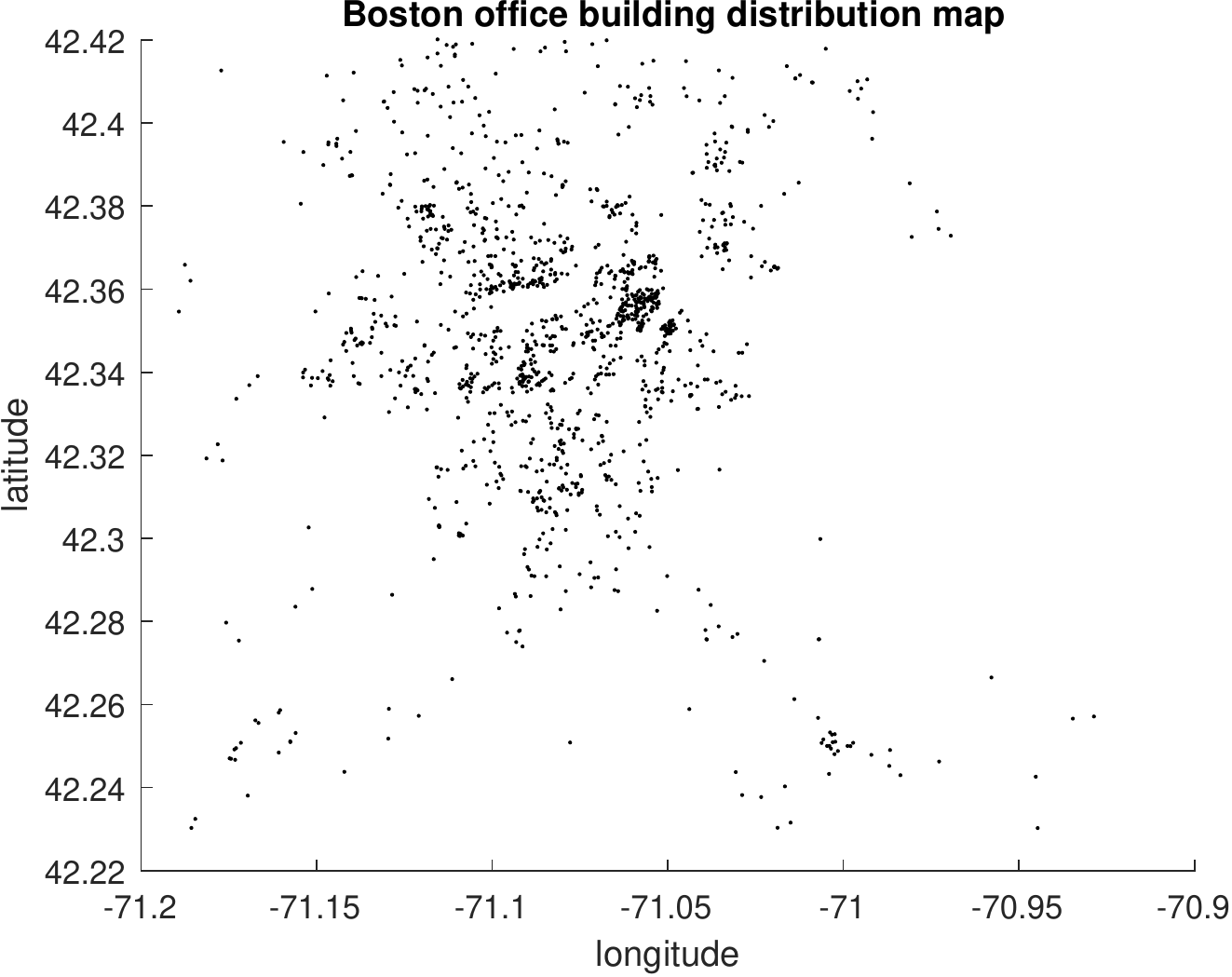}~
		\includegraphics[width=0.3\textwidth]{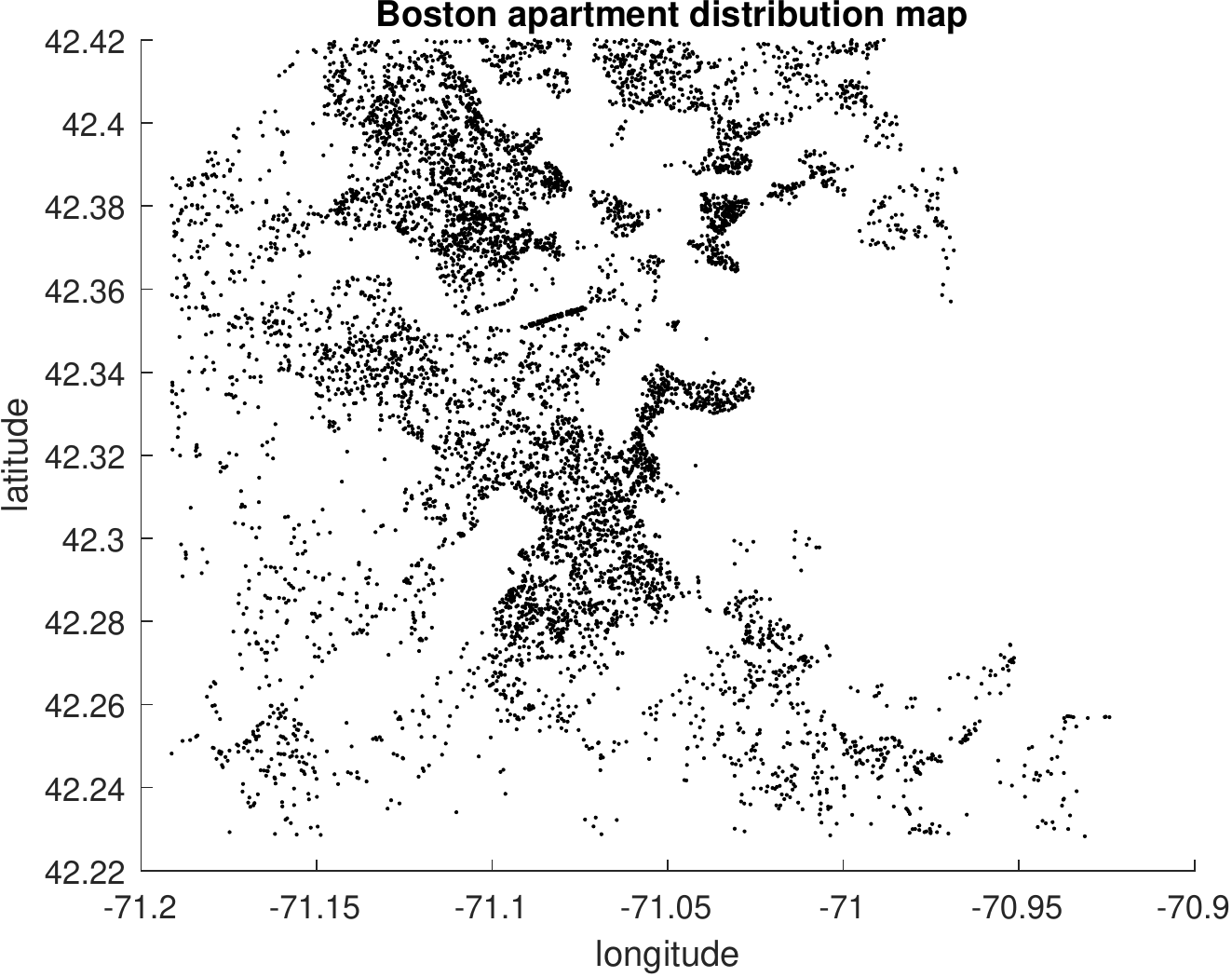}~
		\includegraphics[width=0.3\textwidth]{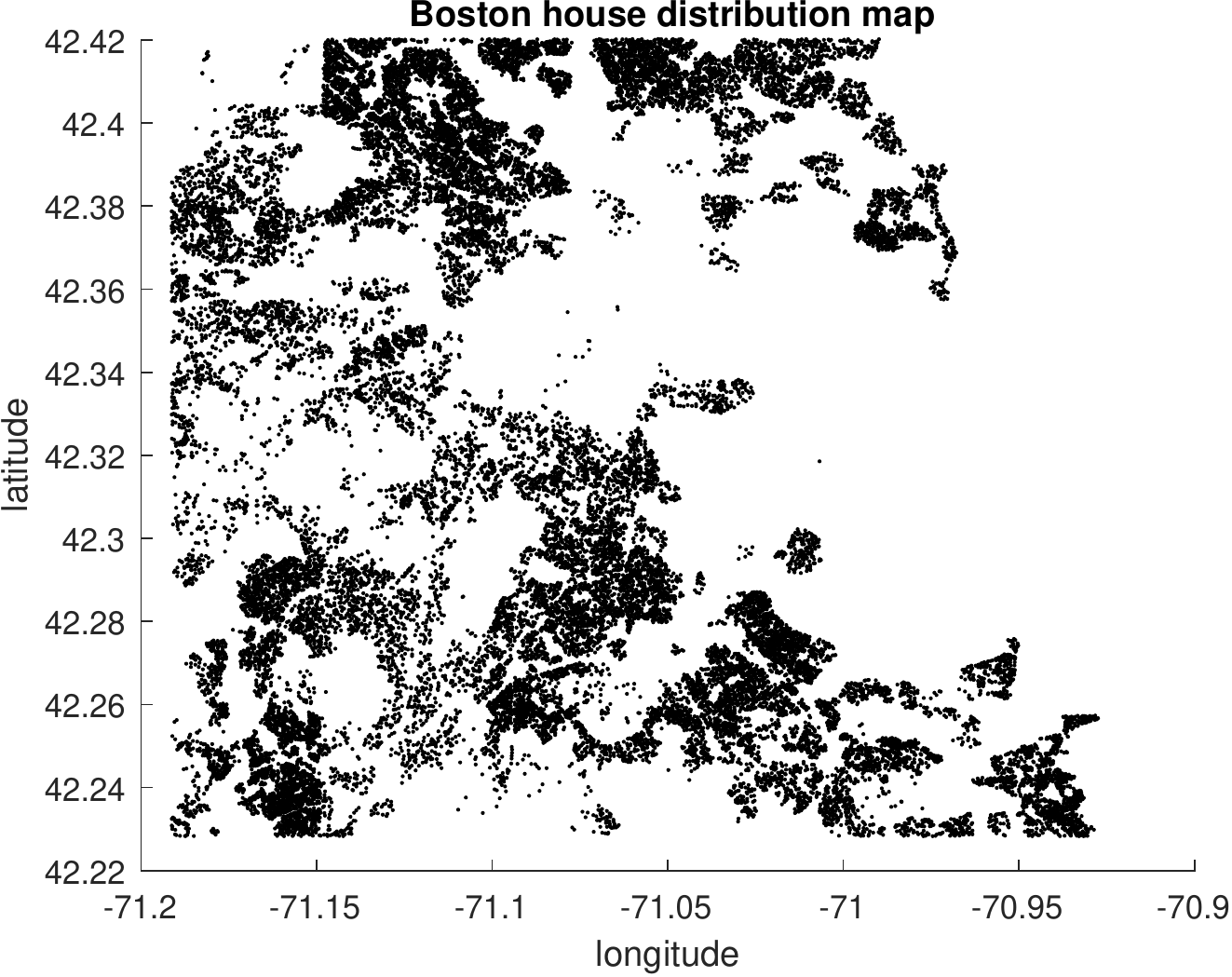}
		\caption{Office building, apartment and house distribution maps of Boston. We can see that both the densities of office buildings and apartments decrease from the center to its outside, while it is contrary of the house density.}
		\label{fg:Boston_office_apartment_house_distribution_maps}
	\end{figure}

	\begin{table}
		\centering
		\caption{Classification performance of randomly selected 1000 buildings of Calgary}
		\begin{tabular}{c|c|c|c|c}
			\hline
			& precision & recall & F1 score & support \\
			\hline
			apartment & 0.54 & 0.77 & 0.64 & 56 \\
			church & 0.00 & 0.00 & 0.00 & 1 \\
			garage & 0.41 & 0.90 & 0.57 & 124 \\
			house & 0.97 & 0.62 & 0.75 & 630 \\
			industrial & 0.51 & 0.80 & 0.63 & 82 \\
			office building & 0.65 & 0.19 & 0.29 & 58 \\
			retail & 0.33 & 0.37 & 0.35 & 43 \\
			roof & 0.15 & 0.83 & 0.25 & 6 \\
			\hline
			\textbf{overall} & 0.78 & 0.64 & 0.66 & 1000 \\
			\hline
		\end{tabular}
		\label{tb:cls_perform_Calgary}
	\end{table}
	According to the map of Boston, houses obviously dominate the buildings of the city, and they are located around the city center. Besides, Boston is also with one single central business district (noted by the black dashed rectangle), since most office buildings and apartments locate in this area. The associated remote sensing image is demonstrated at the bottom of Figure \ref{fg:city_scale_boston_cls_map}. As shown by the distribution maps of office buildings, houses and apartments plotted in Figure \ref{fg:Boston_office_apartment_house_distribution_maps}, we can see that the densities of office buildings and apartments decrease from the center to its outside, while it is contrary of the house density. In addition, Boston is not an industry city, since no large block of industrial districts is observed in the classification map and the proportion of industrial buildings is very low. 
	
\begin{figure}
		\centering
		\includegraphics[width=0.3\textwidth]{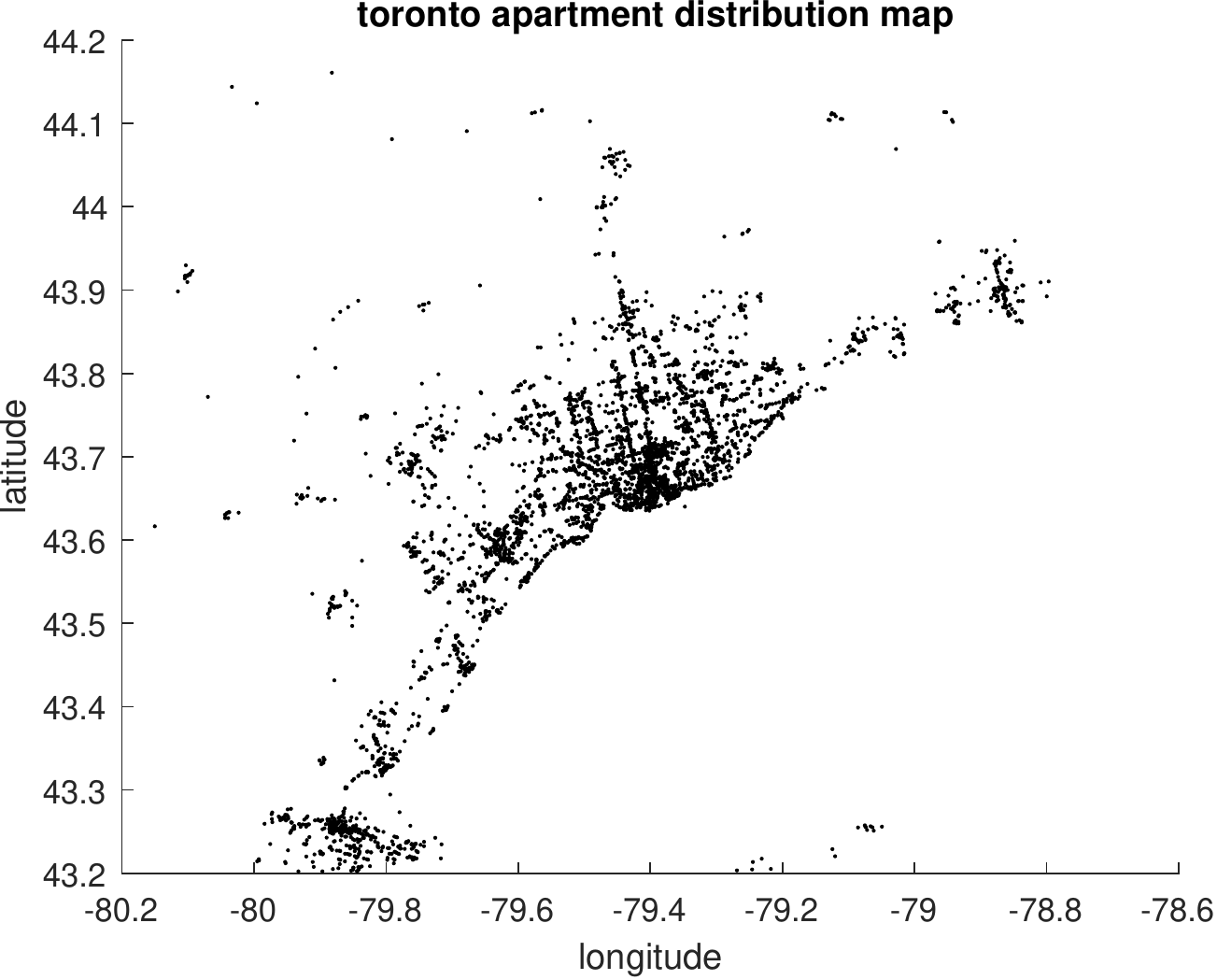}~
		\includegraphics[width=0.3\textwidth]{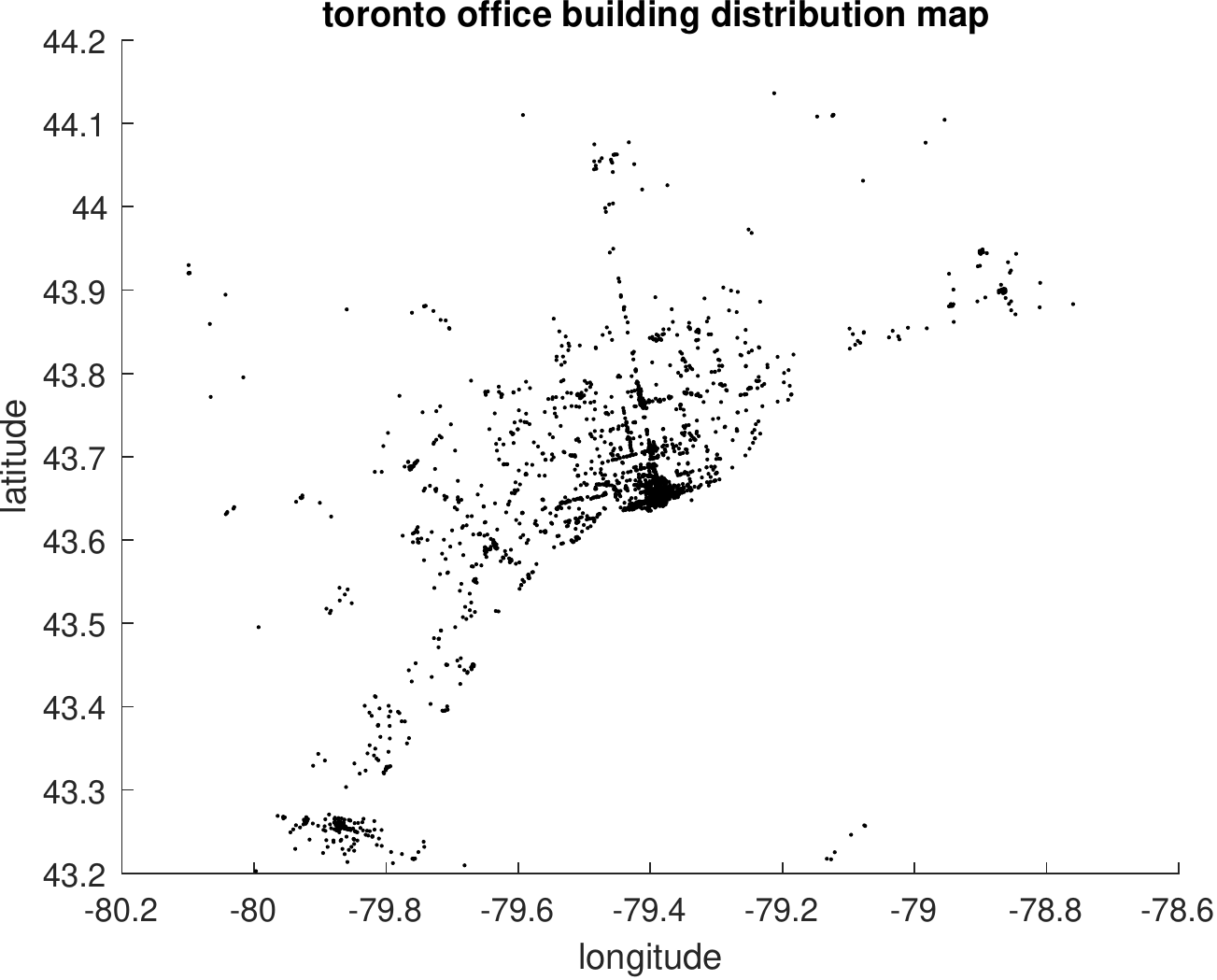}~
		\includegraphics[width=0.3\textwidth]{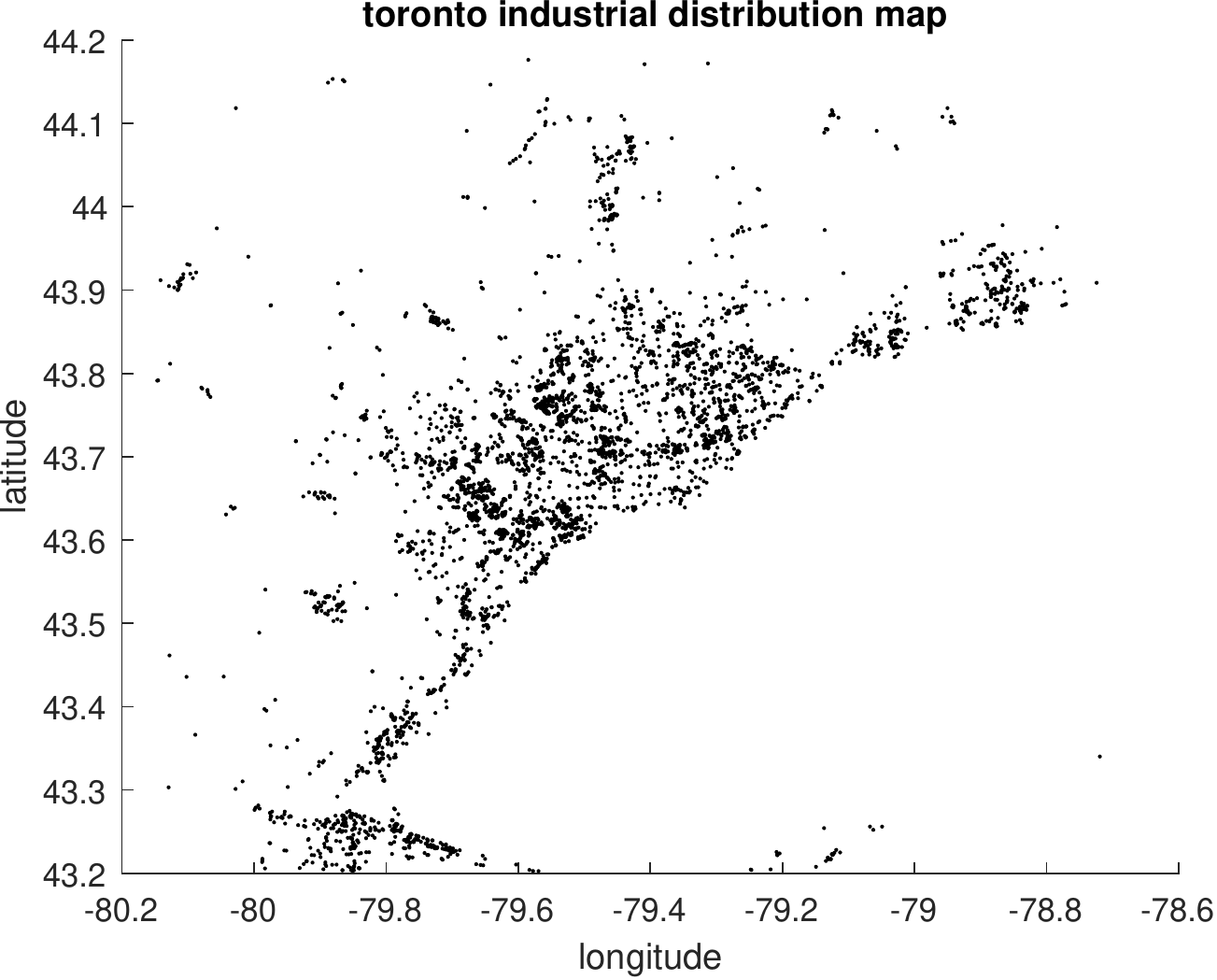}
		\caption{Apartment, office and industrial building distribution maps of Toronto.}
		\label{fg:toronto_building_distribution_maps}
		
	\end{figure}

	From the maps of Figure \ref{fg:toronto_building_cls_map} and \ref{fg:toronto_building_distribution_maps}, most apartments and office buildings are located in the center of Toronto, and most industrial buildings are distributed in the regions around it. As shown in Figure \ref{fg:toronto_pie_chart}, the second largest proportion of building classes is apartment, which indicates that the population density is high in Toronto, especially in the city center. Besides, around $ 10\% $ buildings are industrial, which indicates industry is one of the fields which contribute most to the economy of Toronto.      
	
	\begin{table}
		\centering
		\caption{Classification performance of randomly selected 1000 buildings of Boston}
		\begin{tabular}{c|c|c|c|c}
			\hline
			& precision & recall & F1 score & support \\
			\hline
			apartment & 0.35 & 0.42 & 0.38 & 137 \\
			church & 0.06 & 0.80 & 0.11 & 5 \\
			garage & 0.51 & 0.38 & 0.43  & 221 \\
			house & 0.69 & 0.61 & 0.65 & 546 \\
			industrial & 0.07 & 0.25 & 0.11 & 4 \\
			office building & 0.58 & 0.62 & 0.60 & 60 \\
			retail & 0.20 & 0.42 & 0.27 & 19 \\
			roof & 0.62 & 0.62 & 0.62 & 8 \\
			\hline
			\textbf{overall} & 0.58 & 0.53 & 0.55 & 1000 \\
			\hline
		\end{tabular}
		\label{tb:cls_perform_Boston}
	\end{table}

	
	\begin{table}
		\centering
		\caption{Classification performance of randomly selected 1000 buildings of Toronto}
		\begin{tabular}{c|c|c|c|c}
			\hline
			& precision & recall & F1 score & support \\
			\hline
			apartment & 0.73 & 0.83 & 0.78 & 212 \\
			church & 0.29 & 0.59 & 0.39 & 22 \\
			garage & 0.18 & 0.42 & 0.25  & 33 \\
			house & 0.94 & 0.73 & 0.82 & 575 \\
			industrial & 0.36 & 0.79 & 0.49 & 24 \\
			office building & 0.04 & 0.25 & 0.06 & 4 \\
			retail & 0.84 & 0.50 & 0.63 & 117 \\
			roof & 0.33 & 0.92 & 0.49 & 13 \\
			\hline
			\textbf{overall} & 0.82 & 0.71 & 0.75 & 1000 \\
			\hline
		\end{tabular}
		\label{tb:cls_perform_Toronto}
	\end{table}

	\begin{figure}
		\centering
		\includegraphics[width=0.7\textwidth]{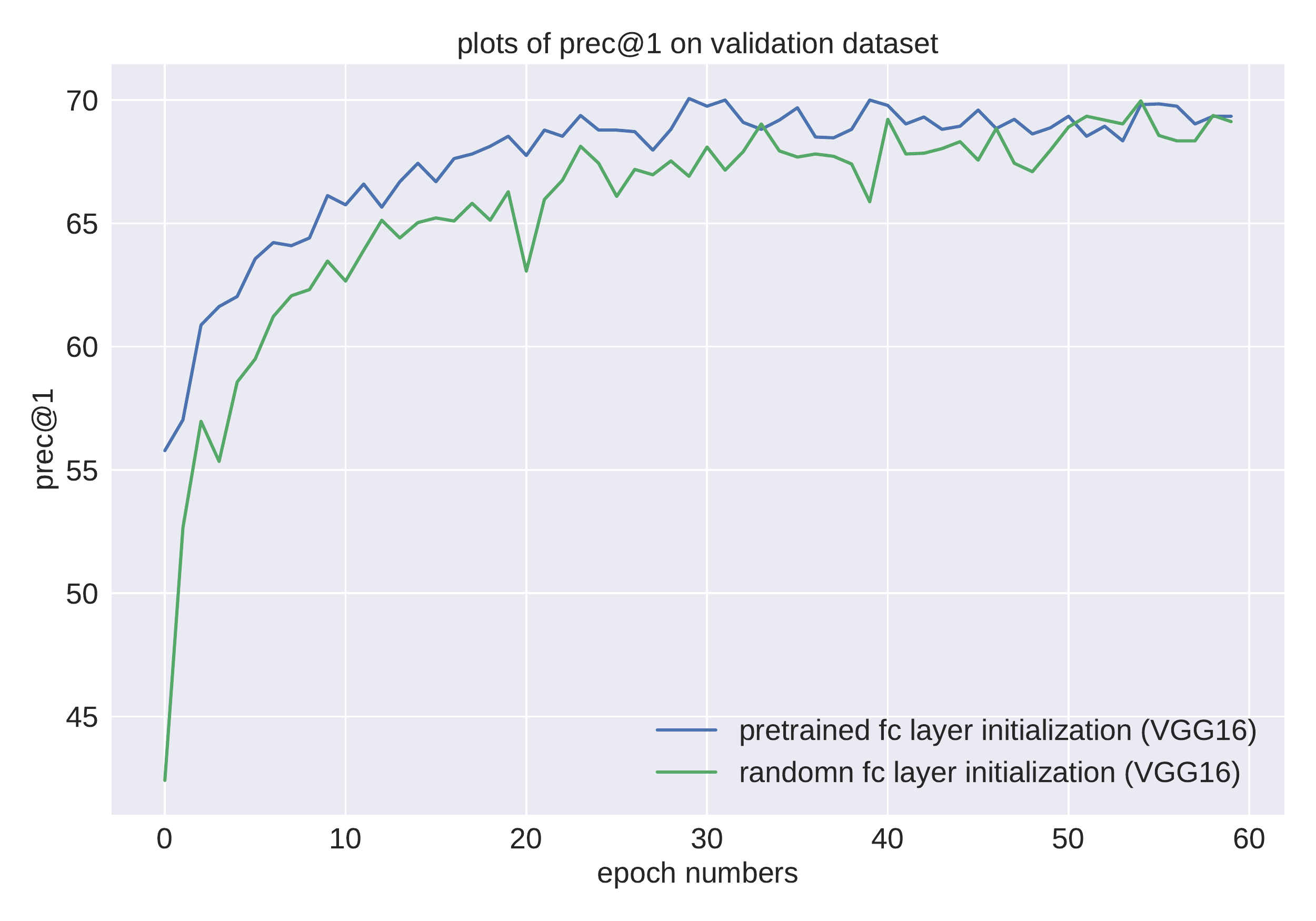}
		\caption{Top-1 precision curves of VGG16 on the validation dataset with different initializations of fully connected layers.}
		\label{fg:vgg16_fc_dif_init}
	\end{figure}
	
	\begin{figure}
		\centering
		\includegraphics[width=0.3\textwidth]{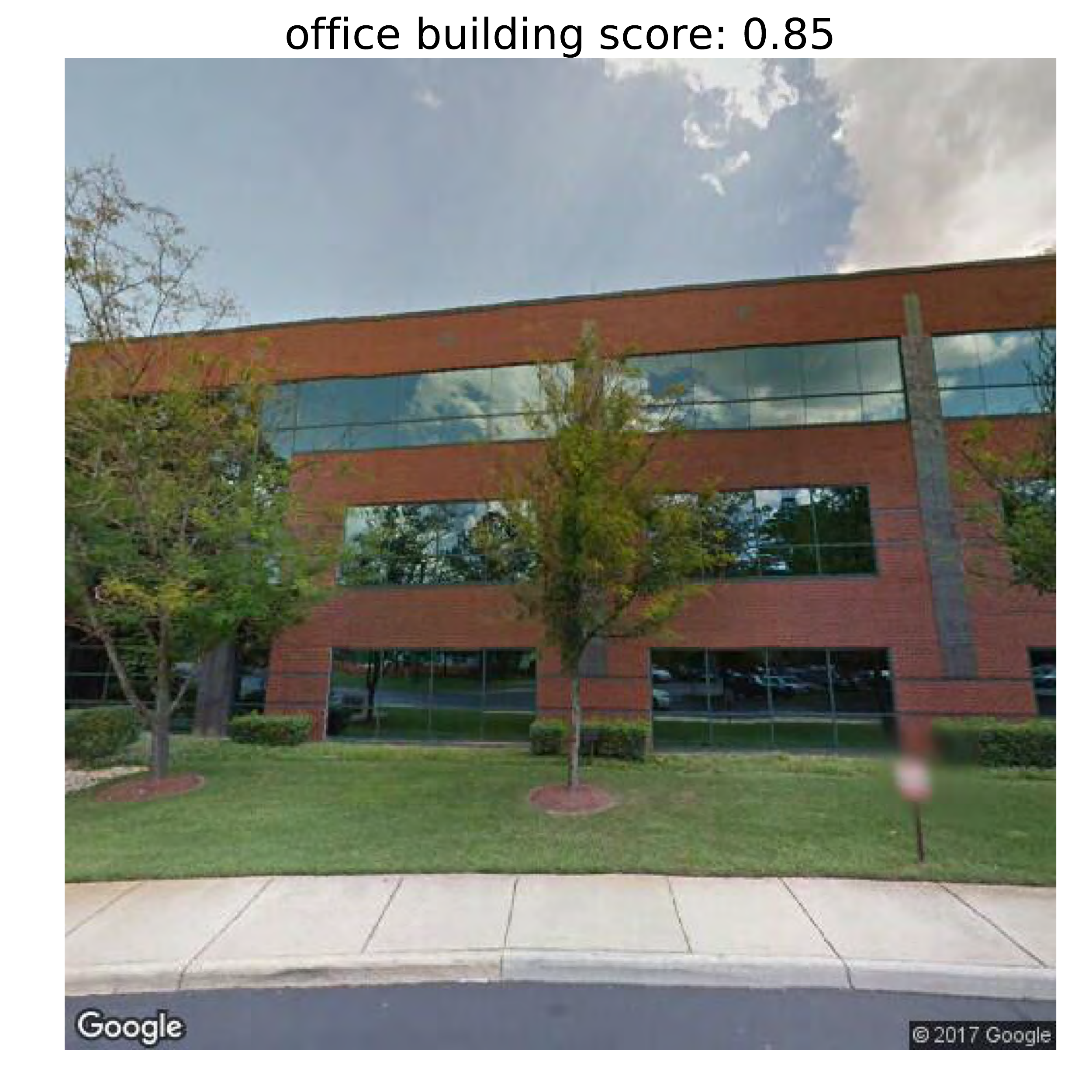}
		~
		\includegraphics[width=0.3\textwidth]{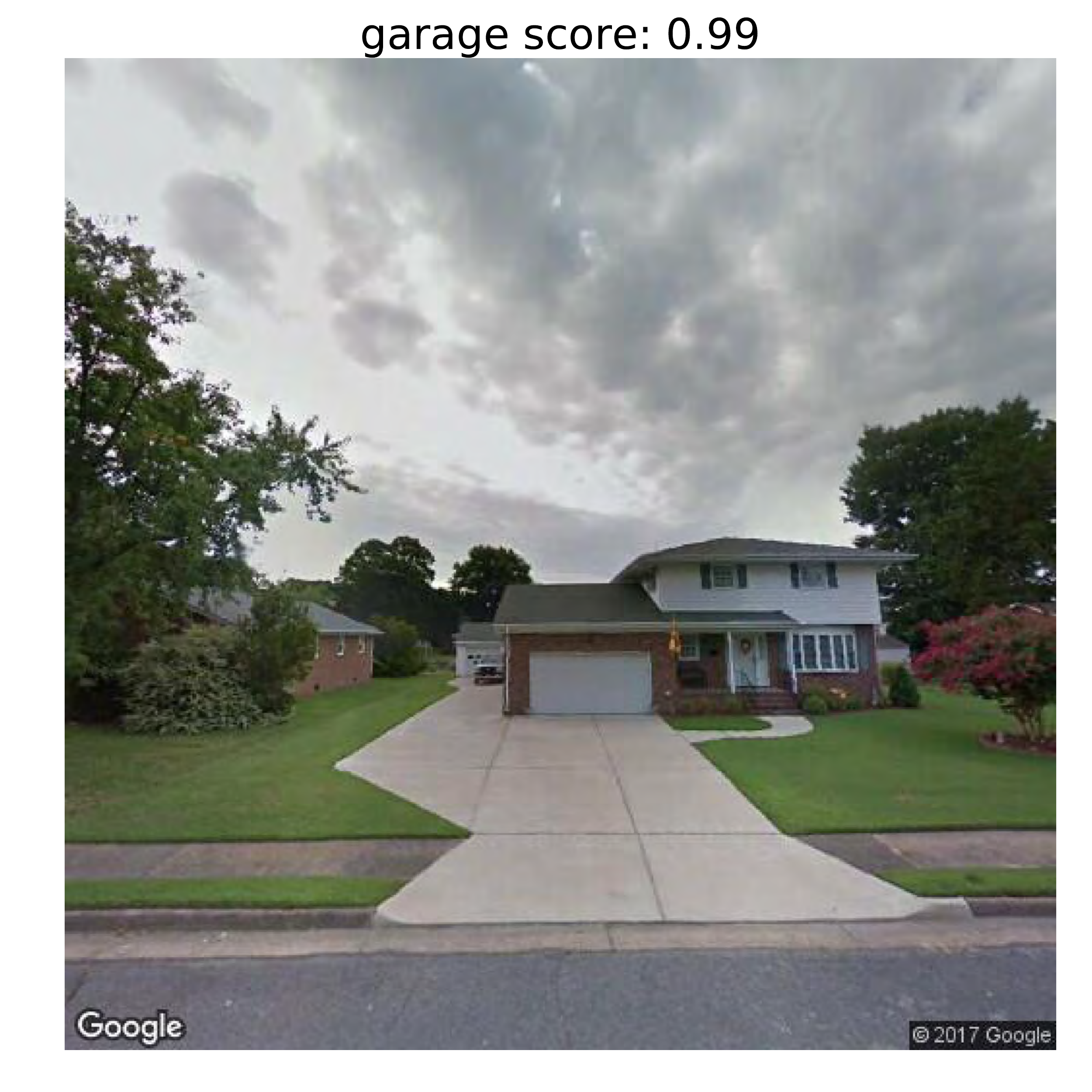}
		~
		\includegraphics[width=0.3\textwidth]{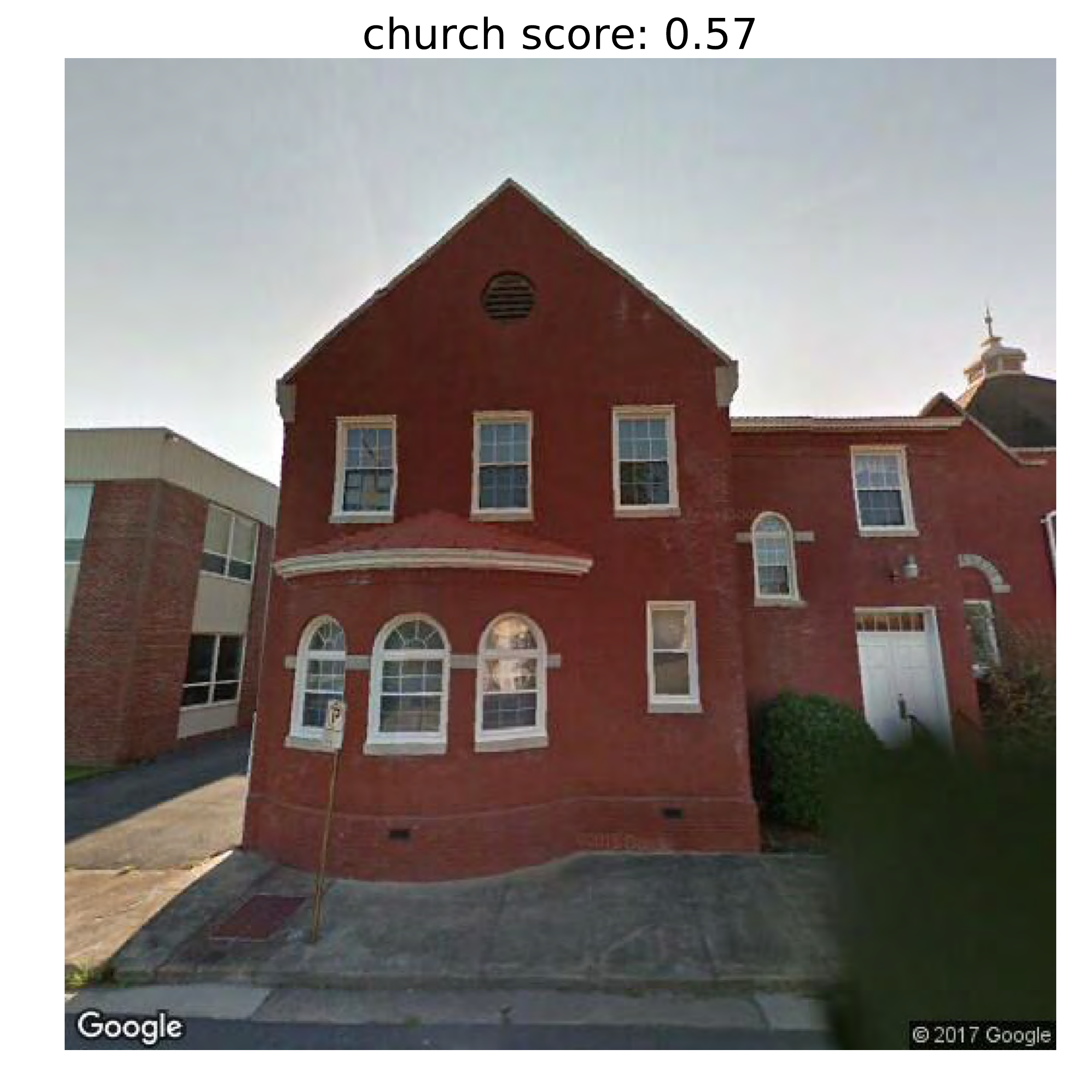}
		\caption{Some results show the reasons that may induce classification errors. (Left) The building displayed by the street view image tends to be an office building, while the ground truth building tag retrieved from OSM is industrial. (Middle) The building demonstrated in the image is a house, while it is misclassified as garage, since there are both garage and house structures demonstrated in the image. (Right) Although the building is correctly recognized, the confidence score is not so high, since the typical fa\c{c}ade structure of church is not displayed in the retrieved image.}
		\label{fg:neg_example_cls_error}
		
	\end{figure}
	
%	\begin{table}
%		\centering
%		\caption{Classification performance of randomly selected 3000 buildings from the three cities}
%		\begin{tabular}{c|c|c|c|c}
%			\hline
%			& precision & recall & F1 score & support \\
%			\hline
%			apartment & 0.50 & 0.62 & 0.55 & 353 \\
%			church & 0.17 & 0.60 & 0.26 & 45 \\
%			garage & 0.38 & 0.53 & 0.44 & 365 \\
%			house & 0.86 & 0.63 & 0.73 & 1816 \\
%			industrial & 0.41 & 0.73 & 0.52 & 112 \\
%			office building & 0.41 & 0.42 & 0.42 & 117 \\
%			retail & 0.48 & 0.46 & 0.47 & 162 \\
%			roof & 0.21 & 0.50 & 0.29 & 30 \\
%			\hline
%			\textbf{overall} & 0.68 & 0.60 & 0.63 & 3000 \\
%			\hline
%		\end{tabular}
%		\label{tb:cls_perform_three_cities}
%	\end{table}

	\section{Discussion}
	In our training experiments, all the fully connected layers are initialized randomly. As for ResNet, there is only one fully connected layer (\textit{softmax} layer) in the architecture and we utilize the network pre-trained on \textit{ImageNet} dataset which contains totally 1000 classes. The parameters of the fully connected layer cannot be directly transferred to our task, since there are 8 classes in our dataset. While, for AlexNet and VGG16, besides the last fully connected layer (\textit{softmax} layer), there are two more fully connected (fc) layers to be initialized. Taking VGG16 as an example, we took two experiments with the same hyperparameters for training the network on our benchmark dataset, while those two fully connected layers were initialized in two ways, i.e. initialized randomly and by the parameters pretrained with \textit{ImageNet}. As shown in \ref{fg:vgg16_fc_dif_init}, VGG16 where the two fully connected layers were initialized by the pretrained parameters did accelerate the training of the network, since it can achieve higher classification accuracy than the one where the fc layers were initialized randomly during the first several epochs. However, both of them can converge to comparable classification accuracies at last.

	According to the classification accuracies of eight building classes, churches are relatively easier to recognize than any other classes, since their structures are more unique, while some classes are not easily identified, e.g. retails and industrials. There are the following reasons that may influence the classification results. Firstly, since the ground truth labels come from the OSM users, manually labeling errors among some building classes exist in the benchmark dataset, especially for those with similar fa\c{c}ade structures, e.g. some industrial and office buildings. As shown in Figure \ref{fg:neg_example_cls_error} (Left), the building displayed by the street view image tends to be an office building, while the building tag retrieved from OSM is industrial. Secondly, some street view images include multiple buildings of different classes, e.g. a house with a garage by its side. From Figure \ref{fg:neg_example_cls_error} (Middle), the building demonstrated in the image is a house, while it is misclassified as a garage. Lastly, side faces of buildings are displayed in some retrieved street view images, thus the corresponding fa\c{c}ade features are not rich for the classification. As illustrated by Figure \ref{fg:neg_example_cls_error} (Right), although the building is correctly recognized, the confidence score is not so high, since the typical fa\c{c}ade structure of church is not displayed in the retrieved image.
	
	It is worth noting that as an alternative, a building rejection class can be added to replace the outlier removal procedure, depending on the quality of input data. 
	
	\section{Conclusion and future work}
	In this paper, we presented a framework for building instance classification, which tended to provide more informative classification maps. With this approach, relatively high accuracies could be achieved for land-use classification of individual buildings. For this task, we built a street view benchmark dataset with eight building categories for training and testing. By investigating four different CNN architectures, we chose VGG16 to predict building instance classification maps on region and city scales. Such maps help us to get insight of urban areas, and have the potential for many innovative urban analysis, e.g. very high resolution urban population density mapping, urban social structure understanding, city economy structure analysis and general urban planning. 
	
	For the future work, to improve the classification performance, other information can be fused, e.g. text descriptions associated with social media images and text information displayed in images, e.g. brand names. Also, in order to obtain denser building classification maps, information from remote sensing images and GIS maps can be exploited for those buildings without street view images. In case that building footprints cannot be retrieved from GIS maps, a method of individual building detection in remote sensing images should be also developed.
	
	\section{Acknowledgment}
	We gratefully acknowledge the support of the European Research Council (ERC) under the European Union’s Horizon 2020 research and innovation programme (grant agreement No [ERC-2016-StG-714087], Acronym: \textit{So2Sat}), Helmholtz Association under the framework of the Young Investigators Group ``SiPEO'' (VH-NG-1018, \url{www.sipeo.bgu.tum.de}), the computing time granted by the John von Neumann Institute for Computing (NIC) and provided on the supercomputer JURECA at Jülich Supercomputing Centre (JSC), as well as NVIDIA Corporation with the donation of the Titan X Pascal GPU used in this research.
	
	The authors would like to thank the reviewers for their valuable suggestions.
	
	
	%% References
	
	
	\bibliographystyle{model1-num-names}
	\bibliography{sample}
	
	
	
	
	
	
	% \end{thebibliography}

\end{document}